\RequirePackage{fix-cm}
\documentclass[smallextended]{svjour3}       
\smartqed  
\usepackage{graphicx}
\usepackage{amsmath}
\usepackage{amssymb}
\usepackage{latexsym}
\usepackage{mathtools}
\usepackage{mathptmx}
\usepackage{nicefrac}
\usepackage{bm}
\usepackage[normalem]{ulem}
\usepackage{tabularx}
\usepackage{caption}
\usepackage{subcaption}

\usepackage[british,english]{babel}

\usepackage{array}
\usepackage{booktabs}

\usepackage{url}
\usepackage{cite}
\usepackage{xcolor}
\definecolor{newcolor}{rgb}{.8,.349,.1}

\usepackage{tikz}
\usepackage{pgfplots}
\pgfplotsset{compat=1.3}
\usepackage{filecontents}
\usetikzlibrary{matrix}
\usetikzlibrary{calc}
\usetikzlibrary{arrows}
\usetikzlibrary{patterns}
\usetikzlibrary{plotmarks}
\usetikzlibrary{intersections} 
\usetikzlibrary{decorations.pathmorphing}
\usetikzlibrary{decorations.pathreplacing}
\usetikzlibrary{decorations.markings,arrows} 
\usetikzlibrary{decorations.shapes}
\usetikzlibrary{fit}
\usetikzlibrary{backgrounds}
\usetikzlibrary{trees}

\usepackage{hyperref}

\usepackage[algoruled,vlined,linesnumbered]{algorithm2e}

\usepackage[colorinlistoftodos]{todonotes}

\journalname{Neural Computing and Applications}

\begin{document}

\title{Character-level Recurrent Neural Networks in Practice: Comparing Training and Sampling Schemes}


\titlerunning{Character-level Recurrent Neural Networks in Practice}        

\author{Cedric De Boom		\and
		Thomas Demeester		\and
		Bart Dhoedt	
}

\authorrunning{C.~De Boom, et al.} 

\institute{Cedric De Boom, Thomas Demeester, Bart Dhoedt \at
              Ghent University, imec, IDLab\\
       			Technologiepark 15\\
       			9052 Ghent, Belgium\\
       			\email{$\{$cedric.deboom, thomas.demeester, bart.dhoedt$\}$@ugent.be}
}


\maketitle

\begin{abstract}
Recurrent neural networks are nowadays successfully used in an abundance of applications, going from text, speech and image processing to recommender systems.
Backpropagation through time is the algorithm that is commonly used to train these networks on specific tasks.
Many deep learning frameworks have their own implementation of training and sampling procedures for recurrent neural networks, while there are in fact multiple other possibilities to choose from and other parameters to tune.
In existing literature this is very often overlooked or ignored.
In this paper we therefore give an overview of possible training and sampling schemes for character-level recurrent neural networks to solve the task of predicting the next token in a given sequence.
We test these different schemes on a variety of datasets, neural network architectures and parameter settings, and formulate a number of take-home recommendations.
The choice of training and sampling scheme turns out to be subject to a number of trade-offs, such as training stability, sampling time, model performance and implementation effort, but is largely independent of the data.
Perhaps the most surprising result is that transferring hidden states for correctly initializing the model on subsequences often leads to unstable training behavior depending on the dataset.

\end{abstract}

\section{Introduction}
Dynamic sequences of discrete tokens are abundant and we encounter them on a daily basis.
Examples of discrete tokens are characters or words in a text, notes in a musical composition, pixels in an image, actions in a reinforcement learning agent, web pages one visits, tracks one listens to on a music streaming service etc.
Each of these tokens appears in a sequence, in which there is often a strong correlation between consecutive or nearby tokens.
For example, the similarity between neighboring pixels in an image is very large since they often share similar shades of colors.
Words in sentences, or characters in words, are also correlated because of the underlying semantics and language characteristics.

In this paper only discrete tokens are considered as opposed to sequences of real-valued samples, such as stock prices, analog audio signals, word embeddings, etc.~but our methodology is also applicable to these kinds of sequences.
A sequence of discrete tokens can be presented to a machine learning model that is designed to assess the probability of the next token in the sequence by modeling $p(x_n | x_{n-1}, x_{n-2}, \dots, x_1)$, in which $x_i$ is the $i$'th token in the sequence.
These kinds of models can go by the names of autoregressive models \cite{Gregor:2013ti}, recurrent or recursive models, dynamical systems etc.
In the field of natural language processing (NLP) they are called language models, in which each token stands for a separate word or $n$-gram \cite{Kim:2015vh}.
Since these models give us a probability distribution of the next token in the sequence, a sample from this distribution can be drawn and thus a new token for the sequence is generated.
By recursively applying this generation step, entire new sequences can be generated.
In NLP, for example, a language model is not only capable of assessing proper language utterances but also of generating new and unseen text.

One particular type of generative models that has become popular in the past years is the recurrent neural network (RNN).
In a regular neural network a fixed-dimensional feature representation is transformed into another feature representation through a non-linear function; multiple instances of such feature transformations are applied to calculate the final output of the neural network.
In a recurrent neural network this process of feature transformations is also repeated in time: at every time step a new input is processed and an output is produced, which makes it suitable for modeling time series, language utterances etc.
These dynamic sequences can be of variable length, and RNNs are able to effectively model semantically rich representations of these sequences.
For example, in 2013, Graves showed that RNNs are capable of generating structured text, such as a Wikipedia article, and even continuous handwriting \cite{Graves:2013ua}.
From then on these models have shown great potential at modeling the temporal dynamics of text, speech as well as audio signals \cite{Karpathy:2015wu, Sercu:2016ub, VanDenOord:2016uo}.
Recurrent neural networks can also effectively generate new images on a per pixel basis, as was shown by Gregor et al.~with DRAW \cite{Gregor:2015up} and by van den Oord et al.~with PixelRNNs \cite{Oord:2016um}.
Next to this, in the context of recommender systems, RNNs have been used successfully to model user behavior on online services and to recommend new items to consume \cite{Tan:2016vy, Hidasi:2015uq}.

Despite the fact that RNNs are abundant in scientific literature and industry, there is not much consensus on how to efficiently train these kinds of models, and, to the extent of our knowledge, there are no focused contributions in literature that tackle this question.
The choice of training algorithm very often depends on the deep learning framework at hand, while in fact there are multiple factors that influence the RNN performance, and those are often ignored or overlooked.
Merity et al.~have pointed out before that ``[the] training of RNN models [...] has fundamental trade-offs that are rarely discussed'' \cite{Merity:2016wg}.
The goal of this paper is to study a number of widely applicable training and sampling techniques for RNNs, along with their respective (dis)advantages and trade-offs.
These will be tested on a variety of datasets, neural network architectures, and parameter settings, in order to gain insights into which algorithm is best suited.
In the next section the concept of RNNs is introduced and, more specifically, character-level RNNs, and how these models are trained.
In Section \ref{sec:schemes} four different training and sampling methods for RNNs are detailed.
After that, in Section \ref{sec:evaluation}, we will present experimental results on the accuracy, efficiency and performance of the different methods.
We will also present a set of take-home recommendations and a range of future research tracks.
Finally, the conclusions are listed in Section \ref{sec:conclusion}.
Table \ref{table:symbols} gives an overview of the symbols that will be used throughout this paper in order of appearance.

\begin{table}[h]
\small
\caption{Tabel of symbols in order of appearance.}
\label{table:symbols}
\begin{tabularx}{\textwidth}{l | X}
\toprule
$x_i$, $\vec{x}_i$	&	Input token $i$, input vector $i$\\
$h_i$, $\vec{h}_i$	&	Hidden state $i$, hidden state vector $i$\\
$y_i$, $\vec{y}_i$	&	Output token $i$, output vector $i$\\
$f(\cdot), g(\cdot)$	&	Parameterized and differentiable functions\\
$\mathcal{L}(\cdot)$	&	Loss function\\
$\eta$				&	Learning rate\\
$w$, $\mathbf{W}$	&	Arbitrary weight, arbitrary weight matrix\\
$k_1$				&	Number of time steps after which one truncated BPTT operation is performed\\
$k_2$				&	Number of time steps over which gradients are backpropagated in truncated BPTT\\
$\phi(\cdot)$		&	Nonlinear activation function\\
$\sigma(\cdot)$		&	Sigmoid function: $\sigma(x) = \nicefrac{1}{\left(1 + \exp(-x)\right)}$\\
$\mathcal{R}$		&	RNN model\\
$\mathcal{R}(\cdot)$	&	Output function of RNN model $\mathcal{R}$\\
$\mathcal{R}_i$		&	Hidden state $i$ of RNN model $\mathcal{R}$\\
$\odot$				&	Element-wise vector multiplication operator\\
$\oplus$				&	Sequence concatenation operator\\
$\mathcal{D}$, $\mathcal{D}_{\text{train}}$, $\mathcal{D}_{\text{test}}$					&	Dataset, train set, test set\\
$V$					&	Ordered set of tokens appearing in a dataset (`vocabulary')\\
$r$					&	Number of recurrent layers in an RNN model\\
$\gamma$				&	Dimensionality of the recurrent layers in an RNN model\\
\bottomrule
\end{tabularx}
\end{table}

\section{Character-level Recurrent Neural Networks}
\label{sec:charrnns}
As mentioned in the introduction, this paper mainly focuses on dynamic sequences of discrete tokens.
Generating and modeling such sequences is the core application of a specific type of recurrent neural networks: character-level recurrent neural networks.
Recurrent neural networks (RNN) are designed to maintain a summary of the past sequence in their memory or so-called \textit{hidden state}, which is updated whenever a new input token is presented.
This summary is used to make a prediction about the next token in the sequence, i.e.~the model $p(x_n | h_{n-1})$ in which the hidden state $h_{n-1}$ of the RNN is a function of the past sequence $(x_{n-1}, x_{n-2}, \dots, x_1)$.
Formally, we have:
\begin{align}
\label{eq:basicrnn}
h_n &= f(x_n, h_{n-1}),\nonumber\\
x_{n+1} &= g(h_n).
\end{align}
Given an adequate initialization $h_0$ of the hidden state and trained parameterized functions $f(\cdot)$ and $g(\cdot)$, the previous scheme can be used to generate an infinite sequence of tokens.
In character-level RNNs specifically, all input tokens are discrete and $g(\cdot)$ is a stochastic function that produces a probability mass function over all possible tokens.
To produce the next token in the sequence, one can sample from this mass function or simply pick the token with the highest probability:
\begin{align}
h_n &= f(x_n, h_{n-1}),\nonumber\\
x_{n+1} &\sim g(h_n).
\end{align}
Since the hidden state at time step $n$ is only dependent on the tokens up to time $n$ and not on future tokens, a character-level RNN can be regarded as the following fully probabilistic generative model \cite{Sutskever:2013wo}:
\begin{align}
\label{eq:generativemodel}
p(x_{1:N}) = \prod_{n=1}^{N} p(x_n | x_{1:n-1}) = \prod_{n=1}^{N} p(x_n, h_{n-1}).
\end{align}
In this, we have used the slice notation $x_{k:\ell}$, which means $(x_k, x_{k+1}, \dots, x_\ell)$.
As a side comment, even though their name only refers to character-based language models, character-level RNNs are fit to model a wide variety of discrete sequences, for which we refer the reader to the introduction section.

\subsection{Truncated backpropagation through time}
\label{sec:TBPTT}
Regular feedforward neural networks are trained using the \textit{backpropagation} algorithm \cite{Rumelhart:1988we}.
In this, a certain input is first propagated through the network to compute the output.
This is called the forward pass.
The output is then compared to a ground truth label using a differentiable loss function.
In the backward pass the gradients of the loss with respect to all the parameters in the network are computed by application of the chain rule.
Finally, all parameters are updated using a gradient-based optimization procedure such as gradient descent \cite{Goodfellow:2016wc}.
In neural network terminology, the parameters of the network are also called the weights.
If the loss function between the network output $y$ and the ground truth label $\hat{y}$ is denoted by $\mathcal{L}(y, \hat{y})$, and the vector of all weights in the network by $\vec{w}$, a standard update rule in gradient descent is given by:
\begin{align}
\label{eq:gradientdescent}
\vec{w} \leftarrow \vec{w} - \eta \cdot \nabla_{\vec{w}} \mathcal{L}(y, \hat{y}).
\end{align}
Here, $\eta$ is the so-called learning rate, which controls the size of the steps taken with each weight update.
In practice, input samples will be organized in equally-sized batches sampled from the training dataset for which the loss is averaged or summed, leading to less noisy updates.
This is called mini-batch gradient descent.
Other gradient descent flavors such as RMSprop and Adam further extend on Equation \eqref{eq:gradientdescent}, making the optimization procedure even more robust \cite{Kingma:2015ku}.

\begin{figure}[h!]
\begin{tikzpicture}[->,thin,>=stealth']
  \tikzstyle{vertex}=[circle,thin,draw=black,minimum size=25pt,inner sep=0pt]
  
  \node[vertex] (x0) at (0,0) {$x_n$};
  \node[vertex] (h0) at (0,1.5) {$h_n$};
  \node[vertex] (x1) at (0,3) {$x_{n+1}$};
  \draw (x0) -- (h0);
  \draw (h0) -- (x1);
  \draw[->,>=stealth'] (h0) to [out=160,in=200,looseness=4] (h0);
  \node (f) at (-0.35,0.8) {$f$};
  \node (g) at (-0.2,2.2) {$g$};
  
  \draw[=>,>=stealth', very thick] (1,1.5) -- (2,1.5);
  \node (unroll) at (1.5,1.8) {unroll};
  
  \node (dots) at (3,1.5) {$\cdots$};
  
  \node[vertex] (x-1) at (4.5,0) {$x_{n-1}$};
  \node[vertex] (h-1) at (4.5,1.5) {$h_{n-1}$};
  \node[vertex] (x-0) at (4.5,3) {$x_{n}$};
  \draw (x-1) -- (h-1);
  \draw (h-1) -- (x-0);
  \draw (dots) -- (h-1);
  \node (f) at (4.0,1.0) {$f$};
  \node (g) at (4.3,2.2) {$g$};
  
  \node[vertex] (x-0) at (6,0) {$x_{n}$};
  \node[vertex] (h-0) at (6,1.5) {$h_{n}$};
  \node[vertex] (x--1) at (6,3) {$x_{n+1}$};
  \draw (x-0) -- (h-0);
  \draw (h-0) -- (x--1);
  \draw (h-1) -- (h-0);
  \node (f) at (5.5,1.0) {$f$};
  \node (g) at (5.8,2.2) {$g$};
  
  \node[vertex] (x--1) at (7.5,0) {$x_{n+1}$};
  \node[vertex] (h--1) at (7.5,1.5) {$h_{n+1}$};
  \node[vertex] (x--2) at (7.5,3) {$x_{n+2}$};
  \draw (x--1) -- (h--1);
  \draw (h--1) -- (x--2);
  \draw (h-0) -- (h--1);
  \node (f) at (7.0,1.0) {$f$};
  \node (g) at (7.3,2.2) {$g$};
  
  \node (dots2) at (9,1.5) {$\cdots$};
  \draw (h--1) -- (dots2);
\end{tikzpicture}
\caption{Unrolling a recurrent neural network in time. Functions $f(\cdot)$ and $g(\cdot)$ and their parameters are shared across all time steps.}
\label{fig:unrolling}
\end{figure}
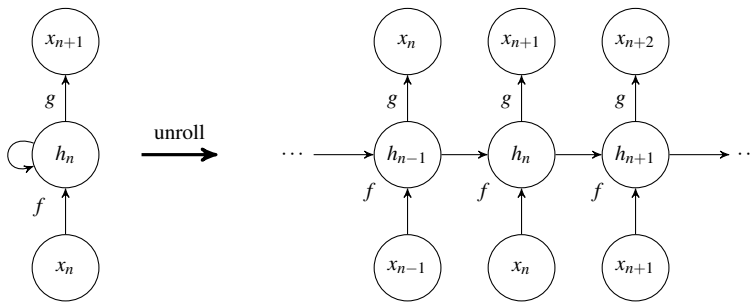

In recurrent neural networks a new input is applied for every time step, and the output at a certain time step is dependent on all previous inputs, as was shown in Equation \eqref{eq:generativemodel}.
This means that the loss at time step $N$ needs to be backpropagated up until the applied inputs at time step $1$.
This procedure is therefore called \textit{backpropagation through time} (BPTT) \cite{Sutskever:2013wo}.
If the sequence is very long, BPTT quickly becomes inefficient: backpropagating through 100 time steps can be compared to backpropagating through a 100-layer deep feedforward neural network.
Unlike with feedforward networks however, in RNNs the weights are shared across time.
This can best be seen if we unroll the RNN to visualize the separate time steps, see Figure \ref{fig:unrolling}.

\begin{figure}[h!]
\begin{tikzpicture}[->,thin,>=stealth']
  \tikzstyle{vertex}=[circle,thin,draw=black,minimum size=25pt,inner sep=0pt]
  
  \node[vertex] (h0) at (0,1.5) {$h_{0}$};
  \node[circle,minimum size=25pt,inner sep=0pt] (xx1) at (0,3) {};
  
  \node[vertex] (x1) at (1.5,0) {$x_{1}$};
  \node[vertex] (h1) at (1.5,1.5) {$h_{1}$};
  \node[vertex] (xx2) at (1.5,3) {$x_{2}$};
  \draw (x1) -- (h1);
  \draw (h1) -- (xx2);
  \draw (h0) -- (h1);
  
  \node[vertex] (x2) at (3,0) {$x_{2}$};
  \node[vertex] (h2) at (3,1.5) {$h_{2}$};
  \node[vertex] (xx3) at (3,3) {$x_{3}$};
  \draw (x2) -- (h2);
  \draw (h2) -- (xx3);
  \draw (h1) -- (h2);
  
  \node[vertex] (x3) at (4.5,0) {$x_{3}$};
  \node[vertex] (h3) at (4.5,1.5) {$h_{3}$};
  \node[vertex] (xx4) at (4.5,3) {$x_{4}$};
  \draw (x3) -- (h3);
  \draw (h3) -- (xx4);
  \draw (h2) -- (h3);
  
  \node[vertex] (x4) at (6,0) {$x_{4}$};
  \node[vertex] (h4) at (6,1.5) {$h_{4}$};
  \node[vertex] (xx5) at (6,3) {$x_{5}$};
  \draw (x4) -- (h4);
  \draw (h4) -- (xx5);
  \draw (h3) -- (h4);
  
  \node[vertex] (x5) at (7.5,0) {$x_{5}$};
  \node[vertex] (h5) at (7.5,1.5) {$h_{5}$};
  \node[vertex] (xx6) at (7.5,3) {$x_{6}$};
  \draw (x5) -- (h5);
  \draw (h5) -- (xx6);
  \draw (h4) -- (h5);
  
  \node[vertex] (x6) at (9,0) {$x_{6}$};
  \node[vertex] (h6) at (9,1.5) {$h_{6}$};
  \node[vertex] (xx7) at (9,3) {$x_{7}$};
  \draw (x6) -- (h6);
  \draw (h6) -- (xx7);
  \draw (h5) -- (h6);
  
  \node[vertex] (x7) at (10.5,0) {$x_{7}$};
  \node[vertex] (h7) at (10.5,1.5) {$h_{7}$};
  \node[vertex] (xx8) at (10.5,3) {$x_{8}$};
  \draw (x7) -- (h7);
  \draw (h7) -- (xx8);
  \draw (h6) -- (h7);
  
  \draw[->, >=stealth', thin] (7.5, -0.7) -- (10.5, -0.7);
  \draw[-, thin] (7.5, -0.75) -- (7.5, -0.65);
  \node (k1) at (9,-0.9) {$k_1$};
  
  \draw[->, >=stealth', thin] (10.5, -1.2) -- (6, -1.2);
  \draw[-, thin] (10.5, -1.25) -- (10.5, -1.15);
  \node (k1) at (8.25,-1.4) {$k_2$};
  
  \draw[->,>=stealth',very thick, shorten >=0.2cm,shorten <=0.2cm] (xx4) to [out=130,in=50] (xx1);
  \draw[->,>=stealth',very thick, shorten >=0.2cm,shorten <=0.2cm] (xx6) to [out=130,in=50] (xx3);
  \draw[->,>=stealth',very thick, shorten >=0.2cm,shorten <=0.2cm] (xx8) to [out=130,in=50] (xx5);
 
\end{tikzpicture}
\caption{Example of trunctated backpropagation through time for $k_1=2$ and $k_2=3$. The thick arrows indicate one backpropagation through time update.}
\label{fig:truncatedbptt}
\end{figure}
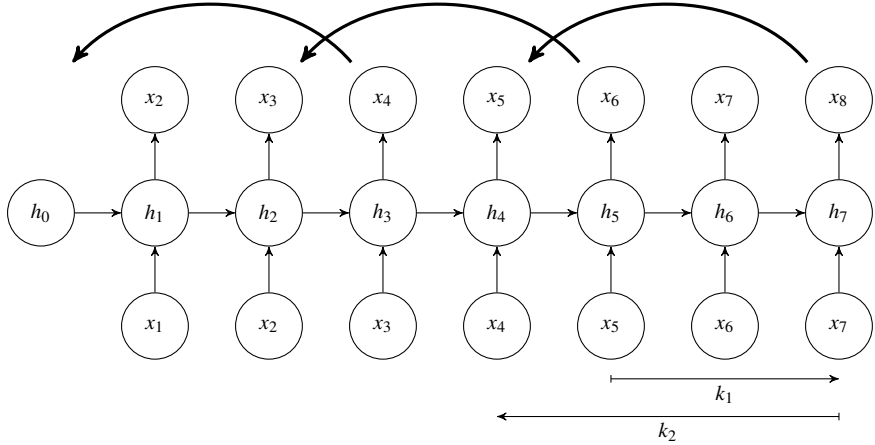

To scale backpropagation through time for use with long sequences, the gradient update is often halted after having traversed a fixed number of time steps.
Such a procedure is called \textit{truncated backpropagation through time}.
Apart from stopping the gradient updates from backpropagating all the way to the beginning of the sequence, we also limit the frequency of such updates.
For a given training sequence, truncated BPTT then proceeds as follows.
Every time step a new token is processed by the RNN, and whenever $k_1$ tokens have been processed in the so-called \textit{forward pass}---and the hidden state is updated $k_1$ times---truncated BPTT is initiated by backpropagating the gradients for $k_2$ time steps.
Here, by analogy with Sutskever \cite{Sutskever:2013wo}, we have denoted the number of time steps between performing truncated BPTT by $k_1$ and the length of the BPTT by $k_2$.
We will keep using these parameters throughout this paper.
A visual explanatory example of truncated BPPT can be found in Figure \ref{fig:truncatedbptt}, which shows how every two ($k_1$) time steps the gradients are backpropagated for three ($k_2$) time steps.
Note that, in order to remain as data efficient as possible, $k_1$ should preferably be less than or equal to $k_2$, since otherwise some data points would be skipped during training.

\subsection{Common RNN layers}
As mentioned before, RNNs keep a summary of the past sequence encoded in a hidden state representation.
Whenever a new input is presented to the RNN, this hidden state gets updated.
The way in which the update happens depends on the internal dynamics of the RNN.
The simplest version of an RNN is an extension of a feedforward neural network, in which matrix multiplications are used to perform input transformations:
\begin{align}
\label{eq:feedforwardlayer}
\vec{u}_{i+1} = \phi\left(\mathbf{W}_i \vec{u}_i + \vec{b}_i\right)
\end{align}
in which $\vec{u}_i$ is the vector representation at the $i$'th layer of the neural network, $\mathbf{W}_i$ is the matrix containing the weights of this layer, and $\vec{b}_i$ is the vector of biases.
The function $\phi(\cdot)$ is a nonlinear transformation function, also called activation function; often used examples are the sigmoid logistic function $\sigma(\cdot)$, $\tanh(\cdot)$ or rectified linear units (ReLU) and their variants.
In the case of RNNs, the hidden state update is rewritten in the style of Equation \eqref{eq:basicrnn} as follows:
\begin{align}
\label{eq:basicrnnlayer}
\vec{h}_n &= \phi\left(\mathbf{W_x}\vec{x}_n + \mathbf{W_h}\vec{h}_{n-1} + \vec{b}\right).
\end{align}
The output of the RNN can be computed with Equation \eqref{eq:feedforwardlayer} using $\vec{h}_n$ as input vector.

To help the RNN model long-term dependencies and to counter the vanishing gradient problem \cite{Hochreiter:2001uj}, several extensions on Equation \eqref{eq:basicrnnlayer} have been proposed.
The best known examples are long short-term memories (LSTMs) and, more recently, gated recurrent units (GRUs), which both have comparable characteristics and similar performance on a variety of tasks \cite{Hochreiter:1997fq, Greff:2015wv, Chung:2014wf}.
Both LSTMs and GRUs incorporate a gating mechanism which controls to what extent the new input is stored in memory and the old memory is forgotten. 
For this purpose the LSTM introduces an input ($\vec{i}_n$) and forget ($\vec{f}_n$) gate, as well as an output gate ($\vec{o}_n$):
\begin{align}
\vec{i}_n &= \sigma\left( \mathbf{U_i}\vec{x}_n + \mathbf{W_i}\vec{h}_{n-1} + \vec{w_i} \odot \vec{c}_{n-1} + \vec{b_i}\right), \nonumber\\
\vec{f}_n &= \sigma\left( \mathbf{U_f}\vec{x}_n + \mathbf{W_f}\vec{h}_{n-1} + \vec{w_f} \odot \vec{c}_{n-1} + \vec{b_f}\right), \nonumber\\
\vec{c}_n &= \vec{f}_t \odot \vec{c}_{n-1} + \vec{i}_n \odot \tanh\left( \mathbf{U_c}\vec{x}_n + \mathbf{W_c}\vec{h}_{n-1} + \vec{b_c} \right), \nonumber\\
\vec{o}_n &= \sigma\left( \mathbf{U_o}\vec{x}_n + \mathbf{W_o}\vec{h}_{n-1} + \vec{w_o} \odot \vec{c}_{n-1} + \vec{b_o}\right), \nonumber\\
\label{eq:lstm}
\vec{h}_n &= \vec{o}_n \odot \tanh\left( \vec{c}_n \right),
\end{align}
Here, the symbol $\odot$ stands for the element-wise vector multiplication.
Note that the LSTM uses two types of memories: a hidden state $\vec{h}_n$ and a so-called cell state $\vec{c}_n$.
Compared to LSTMs, GRUs do not have this extra cell state and only incorporate two gates: a reset and update gate \cite{Cho:2014uo}. This reduces the overall number of parameters and generally speeds up learning.
The choice of LSTMs versus GRUs is dependent on the application at hand, but in the experiments of this paper we will use LSTMs as these are currently the most widely used RNN layers.

Since in neural networks multiple layers are usually stacked onto one another, this is also possible with recurrent layers.
In that case, the output at time step $n$ is fed to the input of the next recurrent layer, also at time step $n$.
Each layer thus processes the sequence of outputs produced by the previous layer.
This will, of course, significantly slow down BPTT.

\section{Training and sampling schemes for character-level RNNs}
\label{sec:schemes}
In this section four schemes are presented on how to train character-level RNNs and how to sample new tokens from them.
The task of the RNN model is independent of these schemes and its purpose is to predict the next symbol or character in a sequence given all previous tokens.
The training and sampling schemes are thus merely a practical means to solve the same task, and in later sections the effect of the used scheme on the performance and efficiency of the RNN model is studied.
We already point out that the four schemes presented are among the most basic and practical methods to train and sample from RNNs, but of course many more combinations or variants could be devised.
In the discussion of the different schemes a general character-level RNN will be denoted by $\mathcal{R}$, and $\mathcal{R}(x)$ is the output of the RNN by applying token $x$ at its input.
This output is a vector that represents a probability distribution across all characters.
For notational convenience, we write the $i$'th hidden state of the RNN by $\mathcal{R}_i$.

\subsection{High-level overview}
As mentioned before, we will isolate four different schemes on how to train RNNs and how to sample new tokens from them.
Each scheme fits in the truncated BPTT framework, and is in fact a practical approximation of the original algorithm.
So whenever we use the $k_1$ and $k_2$ parameters, these refer to the definitions we gave in Section \ref{sec:TBPTT}.
It is also important to keep in mind that the task for all schemes is the same, namely to predict the next token in a given sequence.

To help understand the mechanisms of each scheme, we visualized them schematically in Figures \ref{fig:scheme1}, \ref{fig:scheme2}, \ref{fig:scheme3} and \ref{fig:scheme4}.
In the training procedures we have drawn the output tokens for which a loss is defined.
We see that, for example, the main difference of scheme 2 compared to scheme 1 is that we only compute a loss for the final output token instead of for all output tokens during training.
Regarding the sampling procedures, in the first two schemes a new token is always sampled starting from the same initial hidden state, which is colored light gray.
We call this principle `windowed sampling'.
In schemes 3 and 4, on the other hand, sampling a new token is based on the current hidden state and by applying the previous token at the input of the RNN.
This sampling procedure is called `progressive sampling'.
In the training procedure of scheme 4 we observe a similar technique, in which the hidden state is carried across subsequent sequences.
In the next subsections we will give details on all training and sampling procedures, after which we go over the practical details of the different schemes one by one.
We mention that the schemes are described without batching, while in practice mini-batch training is usually done, as motivated in Section \ref{sec:TBPTT}.
The schemes, however, are easily transferred to a batched setting.


\begin{figure}[h]
\begin{tikzpicture}[->,thin,>=stealth',scale=0.55]
  \tikzstyle{vertex}=[circle,thin,draw=black,minimum size=15pt,inner sep=0pt]
  
  \node[vertex,fill=black!20] (h0) at (0,1.5) {};
  
  \node[vertex] (x1) at (1.5,0) {a};
  \node[vertex] (h1) at (1.5,1.5) {};
  \node[vertex] (xx2) at (1.5,3) {b};
  \draw (x1) -- (h1);
  \draw (h1) -- (xx2);
  \draw (h0) -- (h1);
  
  \node[vertex] (x2) at (3,0) {b};
  \node[vertex] (h2) at (3,1.5) {};
  \node[vertex] (xx3) at (3,3) {c};
  \draw (x2) -- (h2);
  \draw (h2) -- (xx3);
  \draw (h1) -- (h2);
  
  \node[vertex] (x3) at (4.5,0) {c};
  \node[vertex] (h3) at (4.5,1.5) {};
  \node[vertex] (xx4) at (4.5,3) {d};
  \draw (x3) -- (h3);
  \draw (h3) -- (xx4);
  \draw (h2) -- (h3);
  
  \node[vertex,fill=black!20] (h02) at (6.5,1.5) {};
  
  \node[vertex] (x4) at (8,0) {a};
  \node[vertex] (h4) at (8,1.5) {};
  \draw (x4) -- (h4);
  \draw (h02) -- (h4);
  
  \node[vertex] (x5) at (9.5,0) {b};
  \node[vertex] (h5) at (9.5,1.5) {};
  \draw (x5) -- (h5);
  \draw (h4) -- (h5);
  
  \node[vertex] (x6) at (11,0) {c};
  \node[vertex] (h6) at (11,1.5) {};
  \node[vertex] (xx7) at (11,3) {d};
  \draw (x6) -- (h6);
  \draw (h6) -- (xx7);
  \draw (h5) -- (h6);
  
  \node[vertex,fill=black!20] (h03) at (13,1.5) {};
  
  \node[vertex] (x8) at (14.5,0) {b};
  \node[vertex] (h8) at (14.5,1.5) {};
  \draw (x8) -- (h8);
  \draw (h03) -- (h8);
  
  \node[vertex] (x9) at (16,0) {c};
  \node[vertex] (h9) at (16,1.5) {};
  \draw (x9) -- (h9);
  \draw (h8) -- (h9);
  
  \node[vertex] (x10) at (17.5,0) {d};
  \node[vertex] (h10) at (17.5,1.5) {};
  \node[vertex] (xx11) at (17.5,3) {e};
  \draw (x10) -- (h10);
  \draw (h10) -- (xx11);
  \draw (h9) -- (h10);
  
  \node (traintext) at (2.25, 4) {\normalsize Multi-loss training};
  \node (sampletext) at (12, 4) {\normalsize Windowed sampling};
\end{tikzpicture}
\caption{Graphical visualization of scheme 1.}
\label{fig:scheme1}

\vspace*{\floatsep}

\begin{tikzpicture}[->,thin,>=stealth',scale=0.55]
  \tikzstyle{vertex}=[circle,thin,draw=black,minimum size=15pt,inner sep=0pt]
  
  \node[vertex,fill=black!20] (h0) at (0,1.5) {};
  
  \node[vertex] (x1) at (1.5,0) {a};
  \node[vertex] (h1) at (1.5,1.5) {};
  \draw (x1) -- (h1);
  \draw (h0) -- (h1);
  
  \node[vertex] (x2) at (3,0) {b};
  \node[vertex] (h2) at (3,1.5) {};
  \draw (x2) -- (h2);
  \draw (h1) -- (h2);
  
  \node[vertex] (x3) at (4.5,0) {c};
  \node[vertex] (h3) at (4.5,1.5) {};
  \node[vertex] (xx4) at (4.5,3) {d};
  \draw (x3) -- (h3);
  \draw (h3) -- (xx4);
  \draw (h2) -- (h3);
  
  \node[vertex,fill=black!20] (h02) at (6.5,1.5) {};
  
  \node[vertex] (x4) at (8,0) {a};
  \node[vertex] (h4) at (8,1.5) {};
  \draw (x4) -- (h4);
  \draw (h02) -- (h4);
  
  \node[vertex] (x5) at (9.5,0) {b};
  \node[vertex] (h5) at (9.5,1.5) {};
  \draw (x5) -- (h5);
  \draw (h4) -- (h5);
  
  \node[vertex] (x6) at (11,0) {c};
  \node[vertex] (h6) at (11,1.5) {};
  \node[vertex] (xx7) at (11,3) {d};
  \draw (x6) -- (h6);
  \draw (h6) -- (xx7);
  \draw (h5) -- (h6);
  
  \node[vertex,fill=black!20] (h03) at (13,1.5) {};
  
  \node[vertex] (x8) at (14.5,0) {b};
  \node[vertex] (h8) at (14.5,1.5) {};
  \draw (x8) -- (h8);
  \draw (h03) -- (h8);
  
  \node[vertex] (x9) at (16,0) {c};
  \node[vertex] (h9) at (16,1.5) {};
  \draw (x9) -- (h9);
  \draw (h8) -- (h9);
  
  \node[vertex] (x10) at (17.5,0) {d};
  \node[vertex] (h10) at (17.5,1.5) {};
  \node[vertex] (xx11) at (17.5,3) {e};
  \draw (x10) -- (h10);
  \draw (h10) -- (xx11);
  \draw (h9) -- (h10);
  
  \node (traintext) at (2.25, 4) {\normalsize Single-loss training};
  \node (sampletext) at (12, 4) {\normalsize Windowed sampling};
\end{tikzpicture}
\caption{Graphical visualization of scheme 2.}
\label{fig:scheme2}

\vspace*{\floatsep}

\begin{tikzpicture}[->,thin,>=stealth',scale=0.55]
  \tikzstyle{vertex}=[circle,thin,draw=black,minimum size=15pt,inner sep=0pt]
  
  \node[vertex,fill=black!20] (h0) at (0,1.5) {};
  
  \node[vertex] (x1) at (1.5,0) {a};
  \node[vertex] (h1) at (1.5,1.5) {};
  \node[vertex] (xx2) at (1.5,3) {b};
  \draw (x1) -- (h1);
  \draw (h1) -- (xx2);
  \draw (h0) -- (h1);
  
  \node[vertex] (x2) at (3,0) {b};
  \node[vertex] (h2) at (3,1.5) {};
  \node[vertex] (xx3) at (3,3) {c};
  \draw (x2) -- (h2);
  \draw (h2) -- (xx3);
  \draw (h1) -- (h2);
  
  \node[vertex] (x3) at (4.5,0) {c};
  \node[vertex] (h3) at (4.5,1.5) {};
  \node[vertex] (xx4) at (4.5,3) {d};
  \draw (x3) -- (h3);
  \draw (h3) -- (xx4);
  \draw (h2) -- (h3);
  
  \node[vertex,fill=black!20] (h02) at (6.5,1.5) {};
  
  \node[vertex] (x4) at (8,0) {a};
  \node[vertex] (h4) at (8,1.5) {};
  \draw (x4) -- (h4);
  \draw (h02) -- (h4);
  
  \node[vertex] (x5) at (9.5,0) {b};
  \node[vertex] (h5) at (9.5,1.5) {};
  \draw (x5) -- (h5);
  \draw (h4) -- (h5);
  
  \node[vertex] (x6) at (11,0) {c};
  \node[vertex] (h6) at (11,1.5) {};
  \node[vertex] (xx7) at (11,3) {d};
  \draw (x6) -- (h6);
  \draw (h6) -- (xx7);
  \draw (h5) -- (h6);
  
  \node[vertex] (x7) at (12.5,0) {d};
  \node[vertex] (h7) at (12.5,1.5) {};
  \node[vertex] (xx8) at (12.5,3) {e};
  \draw (x7) -- (h7);
  \draw (h7) -- (xx8);
  \draw (h6) -- (h7);
  
  \node[minimum size=15pt] (x8) at (14,0) {$\cdots$};
  \node[minimum size=15pt] (h8) at (14,1.5) {$\cdots$};
  \node[minimum size=15pt] (xx9) at (14,3) {$\cdots$};
  \draw (h7) -- (h8);
  
  \node (traintext) at (2.25, 4) {\normalsize Multi-loss training};
  \node (sampletext) at (10.25, 4) {\normalsize Progressive sampling};
\end{tikzpicture}
\caption{Graphical visualization of scheme 3.}
\label{fig:scheme3}

\vspace*{\floatsep}

\begin{tikzpicture}[->,thin,>=stealth',scale=0.55]
  \tikzstyle{vertex}=[circle,thin,draw=black,minimum size=15pt,inner sep=0pt]
  
  	\tikzoption{diagonal top color}{\pgfsetdiagonaltopcolor{#1}}
	\tikzoption{diagonal bottom color}{\pgfsetdiagonalbottomcolor{#1}}
	\tikzoption{diagonal from left to right}[]{\pgfsetdiagonallefttoright}
	\tikzoption{diagonal from right to left}[]{\pgfsetdiagonalrighttoleft}
	\makeatother
  
  \node[vertex,fill=black!20] (h0) at (0,1.5) {};
  
  \node[vertex] (x1) at (1.5,0) {a};
  \node[vertex] (h1) at (1.5,1.5) {};
  \node[vertex] (xx2) at (1.5,3) {b};
  \draw (x1) -- (h1);
  \draw (h1) -- (xx2);
  \draw (h0) -- (h1);
  
  \node[vertex] (x2) at (3,0) {b};
  \node[vertex,inner color=white,outer color=black!60] (h2) at (3,1.5) {};
  \node[vertex] (xx3) at (3,3) {c};
  \draw (x2) -- (h2);
  \draw (h2) -- (xx3);
  \draw (h1) -- (h2);
  
  \node[vertex] (x3) at (4.5,0) {c};
  \node[vertex] (h3) at (4.5,1.5) {};
  \node[vertex] (xx4) at (4.5,3) {d};
  \draw (x3) -- (h3);
  \draw (h3) -- (xx4);
  \draw (h2) -- (h3);
  
  \node[vertex,inner color=white,outer color=black!60] (h0) at (6,1.5) {};
  \node[vertex] (x1) at (7.5,0) {c};
  \node[vertex] (h1) at (7.5,1.5) {};
  \node[vertex] (xx2) at (7.5,3) {d};
  \draw (x1) -- (h1);
  \draw (h1) -- (xx2);
  \draw (h0) -- (h1);
  
  \node[vertex] (x2) at (9,0) {d};
  \node[vertex] (h2) at (9,1.5) {};
  \node[vertex] (xx3) at (9,3) {e};
  \draw (x2) -- (h2);
  \draw (h2) -- (xx3);
  \draw (h1) -- (h2);
  
  \node[vertex] (x3) at (10.5,0) {e};
  \node[vertex] (h3) at (10.5,1.5) {};
  \node[vertex] (xx4) at (10.5,3) {f};
  \draw (x3) -- (h3);
  \draw (h3) -- (xx4);
  \draw (h2) -- (h3);

  \node[vertex,fill=black!20] (h02) at (12.5,1.5) {};
  
  \node[vertex] (x4) at (14,0) {a};
  \node[vertex] (h4) at (14,1.5) {};
  \draw (x4) -- (h4);
  \draw (h02) -- (h4);
  
  \node[vertex] (x5) at (15.5,0) {b};
  \node[vertex] (h5) at (15.5,1.5) {};
  \draw (x5) -- (h5);
  \draw (h4) -- (h5);
  
  \node[vertex] (x6) at (17,0) {c};
  \node[vertex] (h6) at (17,1.5) {};
  \node[vertex] (xx7) at (17,3) {d};
  \draw (x6) -- (h6);
  \draw (h6) -- (xx7);
  \draw (h5) -- (h6);
  
  \node[vertex] (x7) at (18.5,0) {d};
  \node[vertex] (h7) at (18.5,1.5) {};
  \node[vertex] (xx8) at (18.5,3) {e};
  \draw (x7) -- (h7);
  \draw (h7) -- (xx8);
  \draw (h6) -- (h7);
  
  \node[minimum size=15pt] (x8) at (20,0) {$\cdots$};
  \node[minimum size=15pt] (h8) at (20,1.5) {$\cdots$};
  \node[minimum size=15pt] (xx9) at (20,3) {$\cdots$};
  \draw (h7) -- (h8);
  
  \node (traintext) at (5.25, 4) {\normalsize Conditional multi-loss training};
  \node (sampletext) at (16.25, 4) {\normalsize Progressive sampling};
\end{tikzpicture}
\caption{Graphical visualization of scheme 4. The shaded circle is the remembered hidden state.}
\label{fig:scheme4}
\end{figure}

\subsection{Training algorithms}
We have isolated three different training procedures for character-level RNNs.
A first algorithm is called \textbf{multi-loss training}, and a rudimentary outline of this is shown in Algorithm \ref{algo:sched1train}.
The input sequences all have length $k_2$ and are subsequently taken from the train set by skipping every $k_1$ characters.
For each input token a loss is calculated at the output of the RNN.
When the entire sequence is processed, the average of all losses which we use to update the RNN weights is calculated.
We also reset the hidden state of the RNN for each new training sequence.
This initial hidden state will be learned through backpropagation together with the weights of the RNN.
In practice, for every input sequence of $k_2$ characters, the initial hidden state will be the same.
Note that for LSTMs, the hidden state comprises both the hidden and cell vectors from Equation \eqref{eq:lstm}.
The \texttt{truncated\_BPTT} procedures on lines 10 and 12 calculates the gradients of the loss with respect to all weights in the RNN using backpropagation.
The \texttt{optimize} procedure on lines 11 and 13 then uses these gradients to update the weights using (a variant of) Equation \eqref{eq:gradientdescent}.

\begin{algorithm}[h]
\DontPrintSemicolon
	\SetKwFunction{Init}{initialize}
	\SetKwFunction{TBPTT}{truncated\_BPTT}
	\SetKwFunction{Optimize}{optimize}
	\SetKwData{Loss}{loss}
	\SetKwInOut{Input}{input}
	\SetKwInOut{Output}{output}
	\SetKwInOut{Parameter}{parameters}
	\Input{dataset $\mathcal{D}$ of tokens, RNN $\mathcal{R}$, initial hidden state vector $\vec{h}_0$, loss function $\mathcal{L}$}
	\Parameter{truncated BPTT parameters $k_1$ and $k_2$, learning rate $\eta$}
	\Repeat{\text{convergence}}{
		$j \leftarrow 1$\;
			\While{$j < size(\mathcal{D}) - k_2$}{
				\Loss $\leftarrow 0$\;
				$\mathcal{R}_0 \leftarrow \vec{h}_0$\;
				$s \leftarrow \mathcal{D}_{j:j+k_2}$\;
				\ForEach{token pair $(x_i, x_{i+1}) \in s$}{
					$\vec{y} \leftarrow \mathcal{R}(x_i)$\;
					\Loss $\leftarrow$ \Loss $+\; \nicefrac{\mathcal{L}(\vec{y}, x_{i+1})}{k_2}$\;
				}
				$\forall w \in \mathcal{R}: \nicefrac{\partial \text{\Loss}}{\partial w} \leftarrow$ \TBPTT{$\text{\Loss}, w, k_2$}\;
				$\forall w \in \mathcal{R}:$ \Optimize{$w, \nicefrac{\partial \text{\Loss}}{\partial w}, \eta$}\;
				$\nabla_{\vec{h}_0}\text{\Loss} \leftarrow$ \TBPTT{$\text{\Loss}, \vec{h}_0, k_2$}\;
				\Optimize{$\vec{h}_0, \nabla_{\vec{h}_0}\text{\Loss}, \eta$}\;
				$j \leftarrow j + k_1$
			}
	}
  \caption{Multi-loss training procedure}\label{algo:sched1train}
\end{algorithm}

In \textbf{single-loss training}, instead of defining a loss on all outputs of the RNN---which forces the RNN to make good predictions for the first few tokens of the sequence---we only define a loss on the final predicted token in the sequence.
The complete training procedure is shown in Algorithm \ref{algo:sched2train}.
The difference with Algorithm \ref{algo:sched1train} is that the inner most loop does not aggregate the loss for every RNN output.
Now the loss is calculated outside this loop on line 9, only for the final RNN output.

\begin{algorithm}[h]
\DontPrintSemicolon
	\SetKwFunction{Init}{initialize}
	\SetKwFunction{TBPTT}{truncated\_BPTT}
	\SetKwFunction{Optimize}{optimize}
	\SetKwData{Loss}{loss}
	\SetKwInOut{Input}{input}
	\SetKwInOut{Output}{output}
	\SetKwInOut{Parameter}{parameters}
	\Input{dataset $\mathcal{D}$ of tokens, RNN $\mathcal{R}$, initial hidden state vector $\vec{h}_0$, loss function $\mathcal{L}$}
	\Parameter{truncated BPTT parameters $k_1$ and $k_2$, learning rate $\eta$}
	\Repeat{\text{convergence}}{
		$j \leftarrow 1$\;
			\While{$j < size(\mathcal{D}) - k_2$}{
				$\mathcal{R}_0 \leftarrow \vec{h}_0$\;
				$s \leftarrow \mathcal{D}_{j:j+k_2}$\;
				\ForEach{token pair $(x_i, x_{i+1}) \in s$}{
					$\vec{y} \leftarrow \mathcal{R}(x_i)$\;
					$y' \leftarrow x_{i+1}$\;
				}
				\Loss $\leftarrow \mathcal{L}(\vec{y}, y')$\;
				$\forall w \in \mathcal{R}: \nicefrac{\partial \text{\Loss}}{\partial w} \leftarrow$ \TBPTT{$\text{\Loss}, w, k_2$}\;
				$\forall w \in \mathcal{R}:$ \Optimize{$w, \nicefrac{\partial \text{\Loss}}{\partial w}, \eta$}\;
				$\nabla_{\vec{h}_0}\text{\Loss} \leftarrow$ \TBPTT{$\text{\Loss}, \vec{h}_0, k_2$}\;
				\Optimize{$\vec{h}_0, \nabla_{\vec{h}_0}\text{\Loss}, \eta$}\;
				$j \leftarrow j + k_1$
			}
	}
  \caption{Single-loss training procedure}\label{algo:sched2train}
\end{algorithm}

In both the multi-loss and single-loss procedures we always start training on a sequence from an initial hidden state that is learned.
In \textbf{conditional multi-loss training}, on the other hand, the multi-loss training procedure is adapted to reuse the hidden state across different sequences.
Such an approach leans much closer to the original truncated BPTT algorithm than when the initial state is always reset.
The outline of the training method is given in Algorithm \ref{algo:sched4train}.
Since we are using truncated BPTT, the procedure requires meticulous bookkeeping of the hidden state at every time step, which can be observed in lines 11--12.
This is especially true when we work in a mini-batch setting where we also need to keep track of how the subsequent batches are constructed.

\begin{algorithm}[h!]
\DontPrintSemicolon
	\SetKwFunction{Init}{initialize}
	\SetKwFunction{TBPTT}{truncated\_BPTT}
	\SetKwFunction{Optimize}{optimize}
	\SetKwData{Loss}{loss}
	\SetKwInOut{Input}{input}
	\SetKwInOut{Output}{output}
	\SetKwInOut{Parameter}{parameters}
	\Input{dataset $\mathcal{D}$ of tokens, RNN $\mathcal{R}$, loss function $\mathcal{L}$}
	\Parameter{truncated BPTT parameters $k_1$ and $k_2$, learning rate $\eta$}
	\Repeat{\text{convergence}}{
		$j \leftarrow 0$\;
		$\vec{h}_0 \leftarrow \vec{0}$\;
			\While{$j < size(\mathcal{D}) - k_2$}{
				\Loss $\leftarrow 0$\;
				$\mathcal{R}_0 \leftarrow \vec{h}_0$\;
				$s \leftarrow \mathcal{D}_{j:j+k_2}$\;
				\ForEach{token pair $(x_i, x_{i+1}) \in s$}{
					$\vec{y} \leftarrow \mathcal{R}(x_i)$\;
					\Loss $\leftarrow$ \Loss $+\; \nicefrac{\mathcal{L}(\vec{y}, x_{i+1})}{k_2}$\;
					\If{$i == k_1 - 1$}{
						$\vec{h}_0 \leftarrow \mathcal{R}_i$\;
					}
				}
				$\forall w \in \mathcal{R}: \nicefrac{\partial \text{\Loss}}{\partial w} \leftarrow$ \TBPTT{$\text{\Loss}, w, k_2$}\;
				$\forall w \in \mathcal{R}:$ \Optimize{$w, \nicefrac{\partial \text{\Loss}}{\partial w}, \eta$}\;
				$j \leftarrow j + k_1$
			}
	}
  \caption{Conditional multi-loss training procedure}\label{algo:sched4train}
\end{algorithm}

\subsection{Sampling algorithms}
Next to the training algorithms we have explained in the previous section, we also discern two different sampling procedures.
These are used to generate new and previously unseen sequences.
Both procedures have in common that sampling is started with a seed sequence of $k_2$ tokens that is fed to the RNN.
This is done in order to appropriately bootstrap the RNN's hidden state.
After the seed sequence has been processed, the two procedures start to differ.

In so-called \textbf{windowed sampling} the next token of the sequence is sampled from the RNN after applying the seed sequence.
This newly sampled token is concatenated at the end of the sequence.
After this, the hidden state of the RNN is reset to its learned representation.
Sampling the next token proceeds in the same way: we take the last $k_2$ tokens from the sequence that have been sampled thus far, we feed them to the RNN, the next token is sampled and appended to the sequence, and the hidden state of the RNN is reset.
The entire windowed sampling procedure is sketched in Algorithm \ref{algo:sched1sample}.
On line 6 of the algorithm, we have used the symbol $\oplus$ to indicate sequence concatenation.

\begin{algorithm}[h]
\DontPrintSemicolon
	\SetKwFunction{Init}{initialize}
	\SetKwFunction{TBPTT}{truncated\_BPTT}
	\SetKwFunction{Optimize}{optimize}
	\SetKwFunction{Sample}{sample}
	\SetKwFunction{GetLastKTokens}{get\_last\_k\_tokens}
	\SetKwData{Loss}{loss}
	\SetKwInOut{Input}{input}
	\SetKwInOut{Output}{output}
	\SetKwInOut{Parameter}{parameters}
	\Input{RNN $\mathcal{R}$, initial hidden state vector $\vec{h}_0$, seed sequence $s$}
	\Parameter{truncated BPTT parameter $k_2$}
	\Repeat{\text{enough tokens in} $s$}{
		$\mathcal{R}_0 \leftarrow \vec{h}_0$\;
		$s' \leftarrow$ \GetLastKTokens{$s, k=k_2$}\;
		\ForEach{\text{token} $x \in s'$}{
			$\vec{y} \leftarrow \mathcal{R}(x)$\;
		}
		$s \leftarrow s\; \oplus\; $\Sample{$\vec{y}$}\;
	}
  \caption{Windowed sampling procedure}\label{algo:sched1sample}
\end{algorithm}

In \textbf{progressive sampling} the next token in a sequence is always sampled given the current hidden state and the previously sampled token.
That is, a token is applied at the input of the RNN, which updates its hidden state, and then the next token is sampled at the RNN output.
The initial hidden state is therefore never reset.
This is the most intuitive way of sampling from an RNN.
The entire sampling procedure is given in Algorithm \ref{algo:sched3sample}.
On lines 1--3 the RNN is bootstrapped using the initial hidden state and the seed sequence.
On the following lines, new tokens are continuously sampled from the RNN one token at a time, so the inner loop from Algorithm \ref{algo:sched1sample} is no longer needed.

\begin{algorithm}[h]
\DontPrintSemicolon
	\SetKwFunction{Init}{initialize}
	\SetKwFunction{TBPTT}{truncated\_BPTT}
	\SetKwFunction{Optimize}{optimize}
	\SetKwFunction{Sample}{sample}
	\SetKwFunction{GetLastKTokens}{get\_last\_k\_tokens}
	\SetKwData{Loss}{loss}
	\SetKwInOut{Input}{input}
	\SetKwInOut{Output}{output}
	\SetKwInOut{Parameter}{parameters}
	\Input{RNN $\mathcal{R}$, initial hidden state vector $\vec{h}_0$, seed sequence $s$}
	\Parameter{truncated BPTT parameter $k_2$}
	$\mathcal{R}_0 \leftarrow \vec{h}_0$\;
	\ForEach{\text{token} $x \in s$}{
			$\vec{y} \leftarrow \mathcal{R}(x)$\;
		}
	\Repeat{\text{enough tokens in} $s$}{
		$t \leftarrow\; $\Sample{$\vec{y}$}\;
		$s \leftarrow s \oplus t$\;
		$\vec{y} \leftarrow \mathcal{R}(t)$\;
	}
  \caption{Progressive sampling procedure}\label{algo:sched3sample}
\end{algorithm}

\subsection{Scheme 1 -- Multi-loss training, windowed sampling}
In a first scheme, multi-loss training (Algorithm \ref{algo:sched1train}) is combined with a windowed sampling procedure (Algorithm \ref{algo:sched1sample}).
The main advantage of using this scheme is that there is no need for hidden state bookkeeping across sequences.
Especially during training this can be cumbersome in a batched version of the algorithm.
One disadvantage is that sampling is slower if $k_2$ increases: to sample one new token $k_2$ inputs need to be processed.
If the RNN model contains many layers, this can lead to scalability issues.
Another disadvantage is that a loss is defined on all $k_2$ outputs of the RNN during training.
That is, we force the RNN to produce good token candidates after having seen only one or a few input tokens.
This can lead to a short-sighted RNN model that mostly looks at the more recent history to make a prediction for the next token.
In scheme 2 this potential issue is solved using single-loss training.

\subsection{Scheme 2 -- Single-loss training, windowed sampling}
In the second scheme, the multi-loss training procedure of scheme 1 is replaced by the single-loss equivalent of Algorithm \ref{algo:sched2train}.
The main advantage is that we allow the hidden state of the RNN a certain burn-in period, so that predictions can be made using more knowledge from the past sequence.
Burning in the hidden state also causes the RNN to be able to learn long-term dependencies in the data, because we only make a prediction after having seen $k_2$ tokens.
The potential drawback is that learning is slower, since only one signal is backpropagated for every sequence compared to $k_2$ signals in the first scheme.
The sampling algorithm, on the other hand, is the same as in the first scheme, and now almost perfectly reflects how the RNN has been trained, i.e.~by only considering the final token for each input sequence.

\subsection{Scheme 3 -- Multi-loss training, progressive sampling}
In scheme number 3, we go back to the multi-loss training procedure of scheme 1, but now the progressive sampling from Algorithm \ref{algo:sched3sample} is used instead of windowed sampling.
One drawback of the sampling method in scheme 1 is that it is not very scalable for large values of $k_2$, since we need to feed a sequence of $k_2$ tokens to the RNN for every token that is sampled.
In progressive sampling, on the other hand, the next token is sampled immediately for every new input token.
This way, the sampling of new sequences is sped up by a factor of approximately $k_2$, which is the main advantage of this scheme.

\subsection{Scheme 4 -- Conditional multi-loss training, progressive sampling}
In scheme 3 we still use standard multi-loss training, which resets the hidden state for every train sequence.
Scheme 4 replaces this by the conditional multi-loss training procedure from Algorithm \ref{algo:sched4train}, while maintaining the progressive sampling algorithm.
One of the main disadvantages of using this particular training algorithm, is its requirement to keep track of the hidden states across train sequences and to carefully select these train sequences from the dataset, which can be hard in mini-batch settings.
Next to this, whenever the RNN weights are updated, the hidden state from before the update is reused, which can potentially lead to unstable learning behavior.
On the plus side, we are able to learn dependencies between tokens that are more than $k_2$ time steps away, since the hidden state is remembered in between train sequences.
Also, the need to learn an appropriate initial hidden state is eliminated, which can lead to a small speed-up in learning.

\subsection{Literature overview}
We will now go over some of the works in literature that have used RNNs for language modeling, on both character and word level.
Most of the works that are listed, describe or have described state-of-the-art results on famous benchmarks such as the Penn Treebank dataset \cite{Marcus:1993wd, Mikolov:2010wx}, WikiText-2 \cite{Merity:2016wg} and the Hutter Prize datasets \cite{Hutter:AXQ_crEu}.
The first two datasets are mainly used to benchmark word-level language models, while the Hutter Prize datasets are generally used for character-level evaluation.
Some papers, however, also train character-level models on the Penn Treebank dataset.
It is our purpose to give the reader a high-level idea of what schemes are being used in existing literature.
We do not intend to give a complete overview of the literature on RNNs for language modeling.
Instead, we focus on highly cited works that have, at some point, reported state-of-the-art results on some of the above-mentioned benchmarks.
In this, attention is given to the most recent literature in the field.

The overview can be found in Table \ref{table:literature}.
A distinction is made between character-level models, word-level models and models that are applied on both levels.
At the bottom, three different applications are listed that have used RNNs to model various sequential problems.
We immediately notice that only 5 out of the 22 investigated papers explicitly mention training details regarding loss aggregation or hidden state propagation.
In the other cases we had to go through the source code manually to infer the training and sampling scheme.
If there was no source code available, we contacted the authors directly to ask for more details.
Whenever we could not find information in the paper, the source code or through the authors, we have marked it with `Unknown'.

Scheme number 4 is by far the most popular in recent literature, but scheme number 3 is also widely used.
As mentioned previously, the main difference between these two schemes is whether the transfer of the hidden state between subsequent training sequences takes place or not.
There seems to be no clear consensus on this topic among researchers.
The older works from 2012 and 2013 by Graves \cite{Graves:2013ua} and Mikolov et al.~\cite{Mikolov:2012bw} (and by extension, most of the older works on RNNs) do not transfer the hidden state, while the community seems to be transitioning towards explicitly doing this.
Although there exists no literature describing the advantages and disadvantages of both methods, we can think of some possible explanations for this.
First, while going through multiple source code repositories, we have noticed that source code is often reused by copying and adapting from previous work.
This causes architectural and computational designs to transfer from previous work into other works.
Another possible cause lies with the evolution of deep learning frameworks.
Tensorflow, Keras en PyTorch have made it fairly easy to train RNNs with hidden state transfer, while this was less straightforward or required more effort in older frameworks, such as Theano and Lasagne.

\begin{table}[h]
\small
\caption{Concise literature overview on RNN-based language models.}
\label{table:literature}
\begin{tabular}{l l l l}
\toprule
Reference				& Model type	& Scheme			& Information source\\
\hline
(Graves, 2013) \cite{Graves:2013ua}	& Character-level	& 3	& Author communication\\
(Wu, 2016) \cite{Wu:2016vm}			& Character-level	& Unknown & \\
(Ha, 2016) \cite{Ha:2016ua}			& Character-level	& Unknown & \\
(Cooijmans, 2016) \cite{Cooijmans:2016te}	& Character-level	& 3 & Author communication\\
(Krause, 2016) \cite{BenKrause:2016um}	& Character-level	& 4 & Author communication\\
(Chung, 2016) \cite{Chung:2016tma}		& Character-level	& 4 & Paper\\
(Mujika, 2017) \cite{Mujika:2017uj}		& Character-level	& 4 & Paper\\
\hline
(Zilly, 2017) \cite{Zilly:2017wg}		& Word- \& character-level & 4 & Source code\\
(Melis, 2017) \cite{Melis:2017vx}		& Word- \& character-level & 4 & Paper\\
\hline
(Mikolov, 2012) \cite{Mikolov:2012bw}	& Word-level		& 3		& Source code\\
(Zaremba, 2014) \cite{Zaremba:2014up}	& Word-level		& 4		& Paper\\
(Kim, 2015) \cite{Kim:2015vh}		& Word-level		& 4		& Source code\\
(Gal, 2016) \cite{Gal:2016ti}		& Word-level		& 3		& Source code\\
(Merity, 2016) \cite{Merity:2016wg}		& Word-level		& 2/4 (?)& Author communication\\
(Bradbury, 2016) \cite{Bradbury:2016ul}	& Word-level		& 4 (?)	& Author communication\\
(Zoph, 2016) \cite{Zoph:2016jq}		& Word-level		& 4 		& Author communication\\
(Inan, 2016) \cite{Inan:2016wq}		& Word-level		& 4		& Source code\\
(Merity, 2017) \cite{Merity:2017vl}		& Word-level		& 4		& Source code\\
(Yang, 2017) \cite{Yang:2017ur}		& Word-level		& 3		& Author communication\\
\hline
(Sturm, 2016) \cite{Sturm:2016tv}		& Music notes			& 3 & Author communication\\
(Saon, 2016) \cite{Saon:2016vu}		& Phonemes			& 3 & Author communication\\
(De Boom, 2017) \cite{DeBoom:2017jo}		& Playlist tracks	& 2	& Paper\\
\bottomrule
\end{tabular}
\end{table}

To conclude this concise overview, we have shown that there is a need for clarity and transparency in literature concerning training and sampling details for RNNs.
Not only in the interest of reproducibility, but also to spike awareness in the research community.
This paper is a first attempt at calling attention to the different training and sampling schemes for RNNs, and which trade-offs each of these pose.
In the next section, each of the schemes is evaluated thoroughly in a number of experimental settings.

\section{Evaluation}
\label{sec:evaluation}
In this section, all training and sampling schemes are evaluated in a variety of settings.
As mentioned before, the task in each of these settings is the same: predicting the next token or character in a sequence given the previous tokens or characters.
To perform the evaluation we will use four datasets with different characteristics: English text, Finnish text, C code, and classical music.
Next to this, we will vary the RNN architecture---such as the number of recurrent layers and the hidden state size---as well as the truncated BPTT parameters.
Through these evaluations, we will give some recommendations on how to train and sample from character-level RNNs.

\subsection{Experimental setup}
\label{sec:experimentalsetup}
The central part of our experiments is the RNN model.
For this, we construct a standard architecture with some parameters that we can vary.
The input of the RNN is a one-hot representation of the current character in the sequence, i.e.~a vector of zeros with length $|V|$, with $V$ the ordered set of all characters in the dataset, except for a single one at the current character's position in $V$.
Next, $r$ recurrent LSTM layers are added, each with a hidden state dimensionality of $\gamma$.
In the experiments, the parameters of $r$ and $\gamma$ will be varied.
Finally, we add two fully connected dense layers, one with a fixed dimensionality of 1,024 neurons, and the final dense layer again has dimensionality $|V|$.
At this final layer a softmax function is applied in order to arrive at a probability distribution across all possible next characters.
The complete architecture is summarized in Table \ref{table:architecture} including nonlinear activation functions and extra details.

\begin{table}[h]
\small
\caption{The RNN architecture used in all experiments.}
\label{table:architecture}
\begin{tabular}{l | l}
\toprule
          & Layer type (no.~of dimensions) and nonlinearity\\
\hline
		  & Input ($|V|$)\\
1 to $r$	  & LSTM ($\gamma$)\\
		  & \qquad sigmoid (gates); tanh (hidden and cell state update)\\
		  & \qquad orthogonal initialization, gradient clipping at 50.0\\
$r$ + 1   & Fully connected dense (1,024)\\
          & \qquad leaky ReLU, leakiness $= 0.01$, glorot uniform initialization\\
$r$ + 2   & Fully connected dense ($|V|$)\\
          & \qquad softmax, glorot uniform initialization\\
\bottomrule
\end{tabular}
\end{table}

To train the RNN model we will use one of the schemes outlined in Section \ref{sec:schemes}.
As is common practice in deep learning and gradient-based optimization, multiple training sequences are grouped in batches.
Each sequence in such a batch has a length of $k_2 + 1$ tokens, from which the first $k_2$ are used as input to the RNN, and the next token is used as ground truth signal for every input token.
In this paper, a batch size of 64 sequences is used across all experiments.
To ensure a diverse mix of sequences in each batch, we pick sequences at equidistant offsets, which we increase by $k_1$ for every new batch.
More specifically, every $i$'th batch 64 sequences are sampled at the following offsets in the train set $\mathcal{D}_{\text{train}}$:
\begin{align}
\frac{j \cdot \mathrm{size}(\mathcal{D}_{\text{train}})}{64} + i\cdot k_1,\; j \in \{0, 1, \dots, 63\}.
\end{align}
The entire train set is also circularly shifted after each training epoch.
Since in scheme 4 the hidden states is transferred across different batches, this batching method allows us to fairly compare all four schemes.

At regular points during training the performance of the RNN is evaluated with data from the test set $\mathcal{D}_{\text{test}}$.
For this we will use the perplexity measure, which is widely used in evaluating language models:
\begin{align}
\label{eq:perplexity}
\text{perplexity} = \exp\left( \frac{-\sum_{i=1}^N \log p(x_i | x_{1:i-1})}{N} \right).
\end{align}
In this formula $x_i$ is the $i$'th token in the sequence and $N$ is the total number of tokens.
The better a model is at predicting the next token, the lower its perplexity measure.
In the context of RNNs, the conditional probability in Equation \eqref{eq:perplexity} is approximated using the hidden state of the RNN, as was shown in Equation \eqref{eq:generativemodel}.
In practice, the hidden state of the RNN is bootstrapped with $k_2$ characters and perplexity is calculated on all subsequent characters in the test set.

In the experiments below, every RNN is trained with 12,800 batches of 64 sequences using the batching method described above.
For all schemes and experiments the standard categorical cross-entropy loss function is used, which calculates the inner product between the log output probability vector $\vec{y}$ and the one-hot vector of the target token $\hat{x}$:
\begin{align}
\mathcal{L}\left(\vec{y}, \hat{x}\right) = - \vec{onehot}\!\left(\hat{x}\right) \cdot \log\!\left(\vec{y}\right).
\end{align}
During training we report perplexity on the test set at logarithmically spaced intervals.
All RNN models are trained five times with $k_2$ always set to 100 (unless explicitly indicated otherwise), and we choose $k_1\in\{20, 40, 60, 80, 100\}$.
For every new configuration we reinitialize all network weights and random generators to the same initial values.
As optimization algorithm we use Adam with a learning rate of 0.001 throughout the experiments.

All experiments are performed on a single machine, 12 core Intel Xeon E5-2620 2.40GHz CPU and Nvidia Tesla K40c GPU.
We use a combination of Theano 0.9 and Lasagne 0.2 as implementation tools, powered by cuDNN 5.0.

\subsection{Datasets}
In the experiments the performance of each scheme is evaluated on four datasets\footnote{The datasets are available for download at \url{https://github.com/cedricdeboom/character-level-rnn-datasets}}.
The dataset characteristics of these datasets.
\begin{enumerate}
\item English: we compiled all plays by William Shakespeare from the Project Gutenberg website\footnote{\url{www.gutenberg.org}} in one dataset.
The plays follow each other in random order.
The total number of characters is 6,347,705 with 85 unique characters.
\item Finnish: this language is very different from English. On the Gutenberg website we gathered all texts from Finnish playwrights Juhani Aho and Eino Leino.
This results in a dataset of 10,976,530 characters, of which 106 are unique.
\item Linux: we saved all C code from the Linux kernel\footnote{\url{github.com/torvalds/linux/tree/master/kernel}} and gathered the files together.
On November 22 2016, the entire kernel contained 6,546,665 characters, and 97 of them are unique.
\item Music: we created this dataset by extracting music notes from MIDI files.
When notes are played simultaneously in the MIDI file, we extract them from low to high, so that we obtain a single sequence of subsequent notes.
We downloaded all piano compositions by Bach, Beethoven, Chopin and Haydn from Classical Archives\footnote{\url{www.classicalarchives.com}}, removed duplicate compositions, and gathered a dataset of 1,553,852 notes, of which there are 90 unique ones.
\end{enumerate}
After cyclically permuting each dataset over a randomly chosen offset, we extract the last 11,100 characters to compile a test set $\mathcal{D}_{\text{test}}$.
All remaining characters form the train set $\mathcal{D}_{\text{train}}$.

\subsection{Experiments}
\begin{figure}[h]
\begin{flushleft}
    \begin{subfigure}[b]{0.45\textwidth}
        \includegraphics[trim = 0.5cm 0.2cm 1.5cm 0.5cm, clip, width=\textwidth]{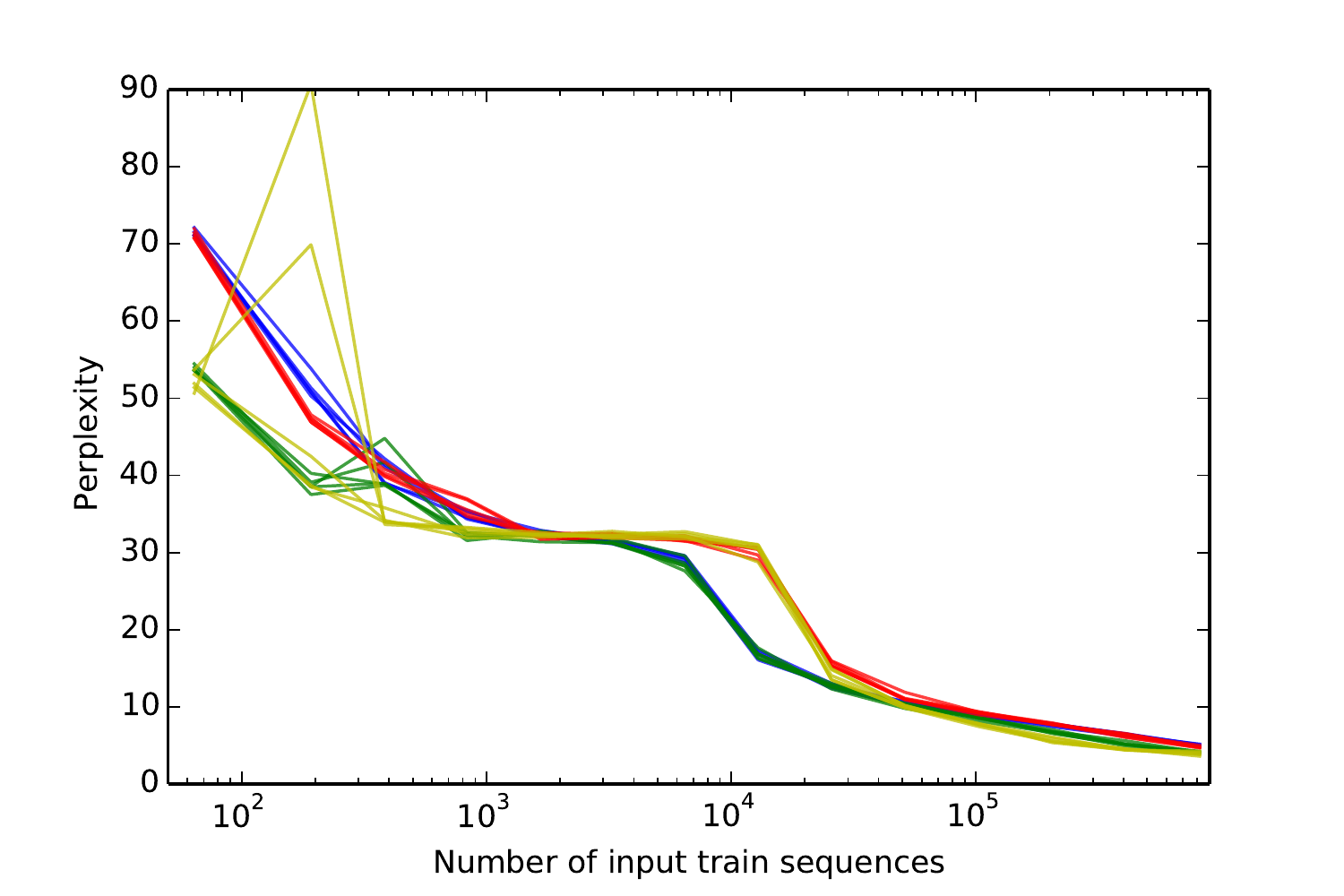}
        \caption{English}
        \label{fig:shakespeare1architectures}
    \end{subfigure}
    \begin{subfigure}[b]{0.45\textwidth}
        \includegraphics[trim = 0.5cm 0.2cm 1.5cm 0.5cm, clip, width=\textwidth]{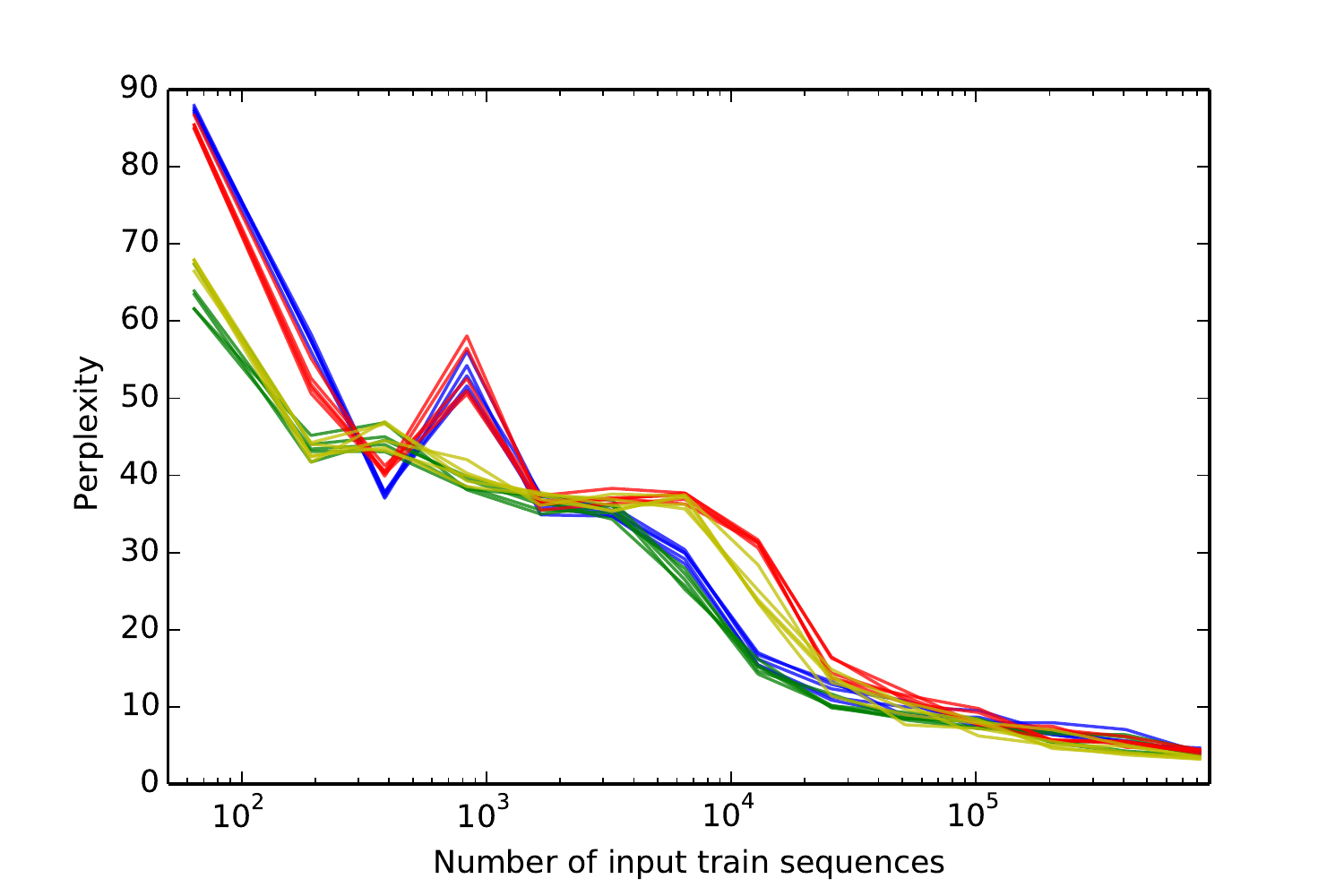}
        \caption{Finnish}
        \label{fig:finnish1architectures}
    \end{subfigure}
    \begin{subfigure}[b]{0.45\textwidth}
        \includegraphics[trim = 0.5cm 0.2cm 1.5cm 0.5cm, clip, width=\textwidth]{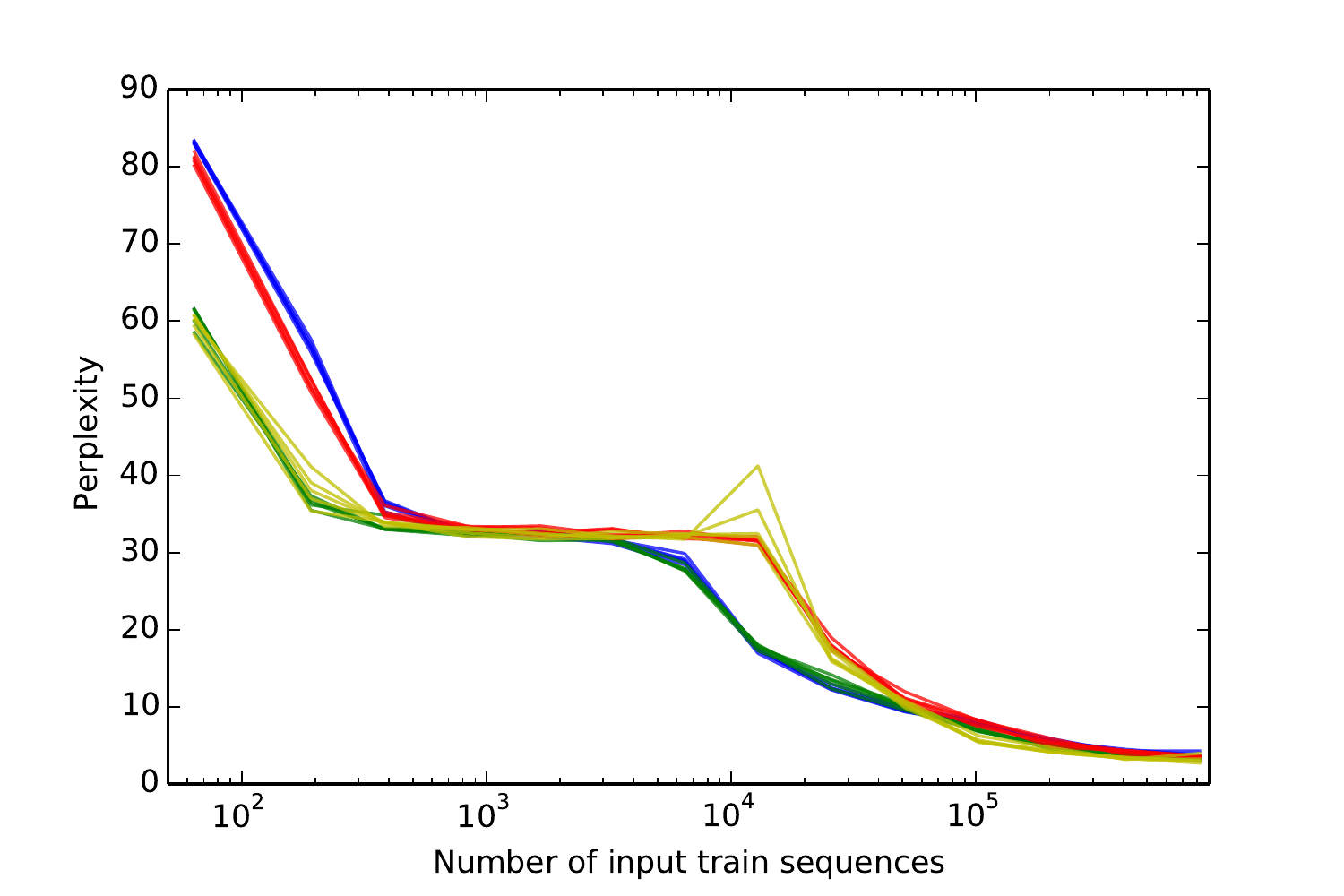}
        \caption{Linux}
        \label{fig:linux1architectures}
    \end{subfigure}
    \begin{subfigure}[b]{0.45\textwidth}
        \includegraphics[trim = 0.5cm 0.2cm 1.5cm 0.5cm, clip, width=\textwidth]{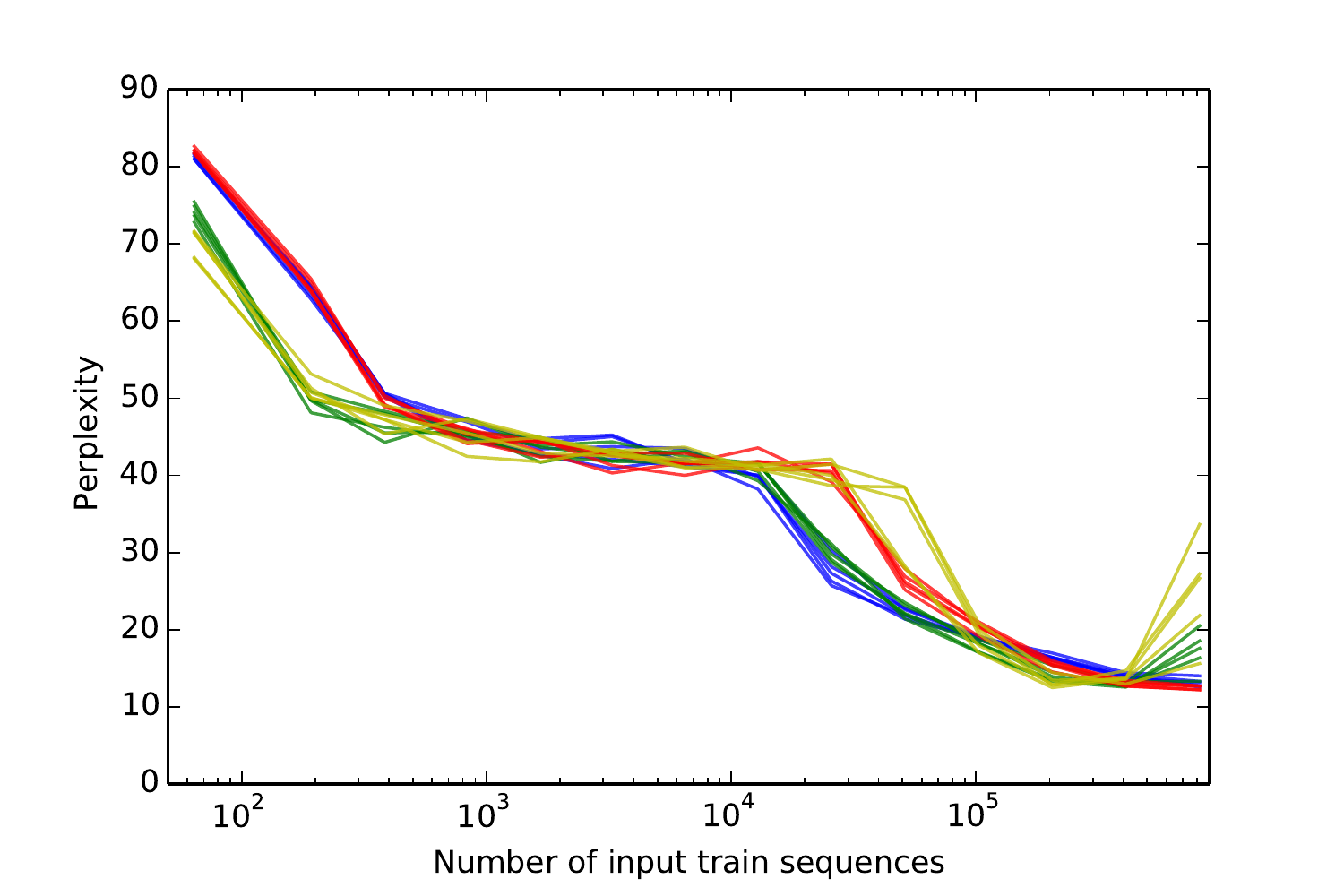}
        \caption{Music}
        \label{fig:music1architectures}
    \end{subfigure}
\end{flushleft}
    \caption{Comparing RNN architectures with scheme 1. Blue: $r=1, \gamma=128$; Green: $r=1, \gamma=512$; Red: $r=2, \gamma=128$; Yellow: $r=2, \gamma=512$.}
    \label{fig:exp1architectures}
\end{figure}

Several experiments are now performed to evaluate the predictive performance of RNNs that have been trained with different configurations.
In a first round of experiments the architecture of the RNN models is varied.
More specifically, we set the number of recurrent LSTM layers $r$ to 1 or 2, and we also change the hidden state size $\gamma$ to either 128 or 512.
Figure \ref{fig:exp1architectures} shows plots for these different RNN architectures, trained using scheme 1 and on all four datasets.
For every architecture we have plotted five lines for the different settings of $k_1$ mentioned above.
In all plots we see that the RNNs with $\gamma=512$ (green and yellow) initially perform better, but the RNNs with $r=1$ (green and blue) learn somewhat faster on the long term.
At 12,800 batches there is no clear difference in performance anymore between the architectures.
On the music dataset and architectures with $\gamma=512$ we observe some overfitting.
If we add $25\%$ of dropout to the final two dense layers \cite{Srivastava:2014ww}, this overfitting is already greatly reduced, but still observable (not shown in the graph).
For the Finnish dataset and architectures with $\gamma=128$ we notice a bump around 1,000 train sequences, which is present for all $k_1$ configurations.
This bump is lowered if we reduce the learning rate to 0.0001 or use a different, non momentum-based optimizer such as RMSProp, but it remains an artefact of both the dataset and architecture.

\begin{figure}[h]
\begin{flushleft}
    \begin{subfigure}[b]{0.45\textwidth}
        \includegraphics[trim = 0.5cm 0.2cm 1.5cm 0.5cm, clip, width=\textwidth]{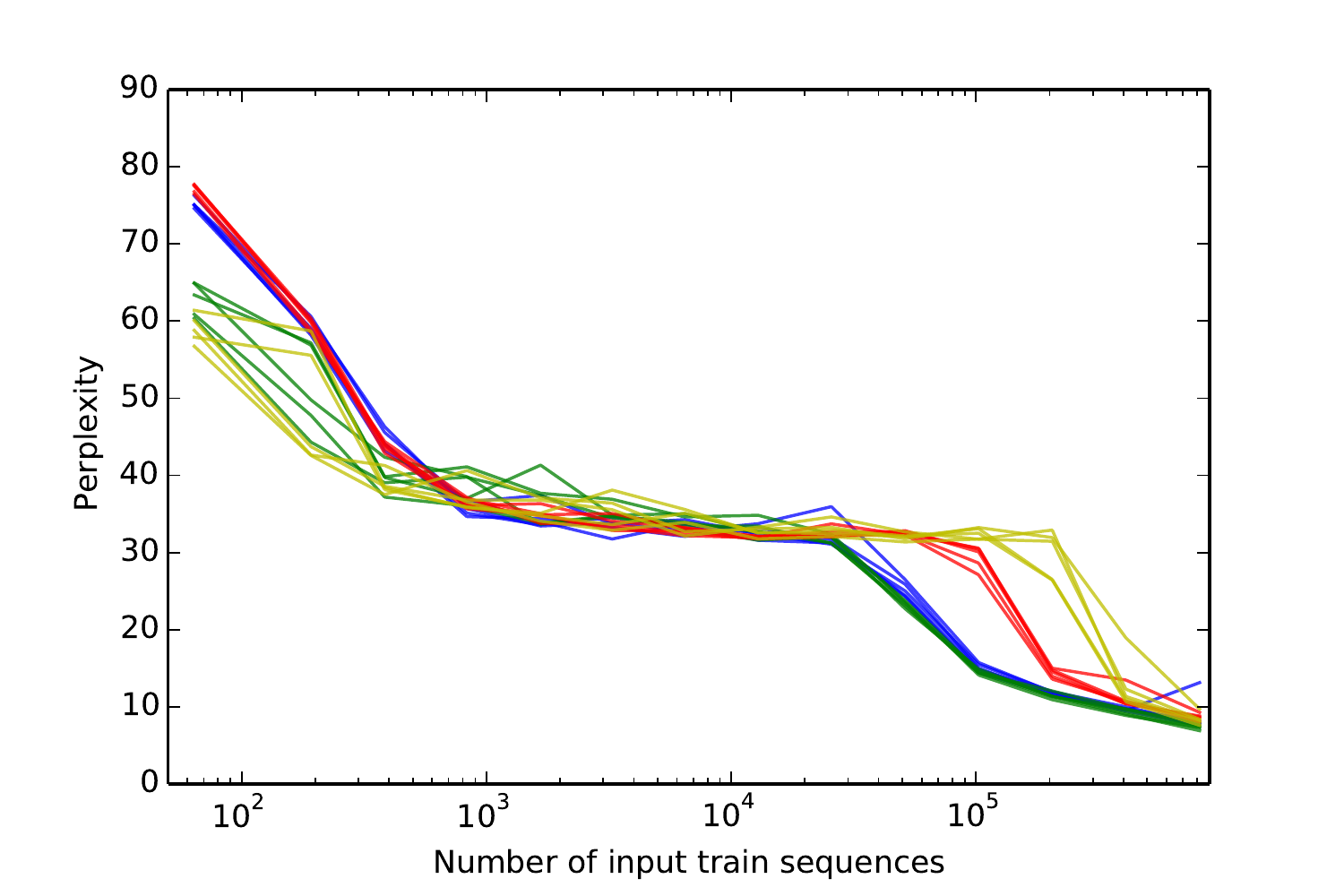}
        \caption{English}
        \label{fig:shakespeare2architectures}
    \end{subfigure}
    \begin{subfigure}[b]{0.45\textwidth}
        \includegraphics[trim = 0.5cm 0.2cm 1.5cm 0.5cm, clip, width=\textwidth]{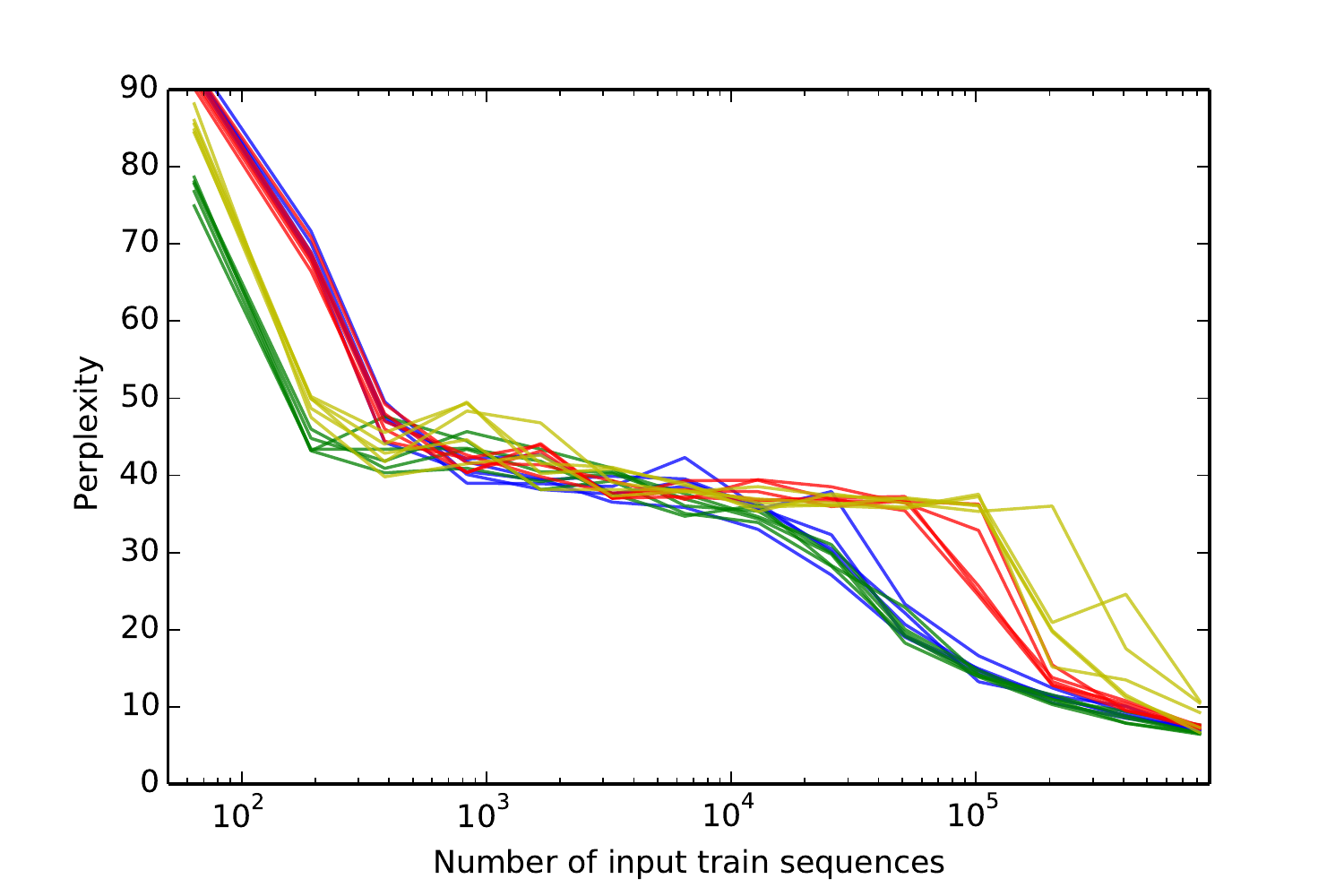}
        \caption{Finnish}
        \label{fig:finnish2architectures}
    \end{subfigure}
    \begin{subfigure}[b]{0.45\textwidth}
        \includegraphics[trim = 0.5cm 0.2cm 1.5cm 0.5cm, clip, width=\textwidth]{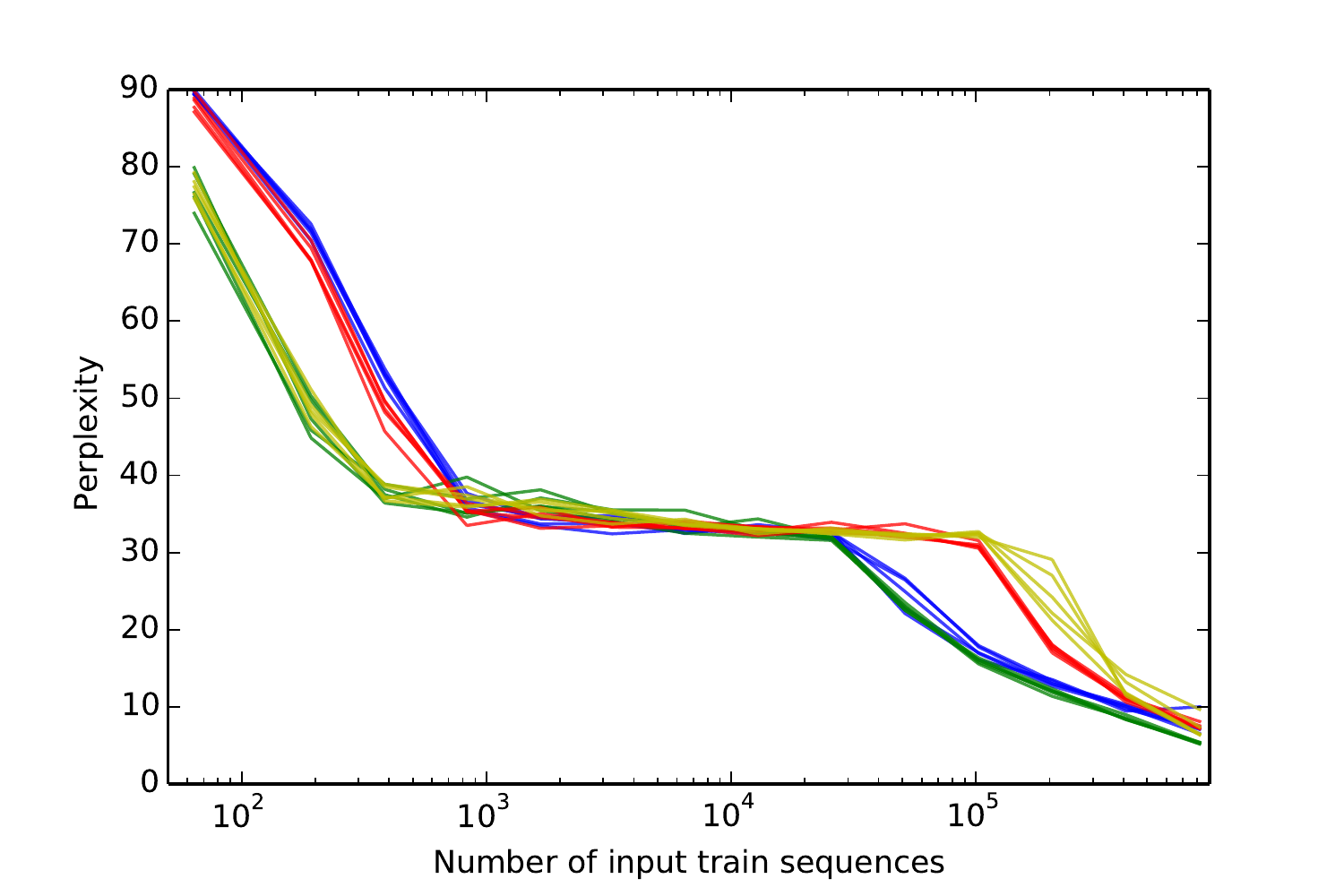}
        \caption{Linux}
        \label{fig:linux2architectures}
    \end{subfigure}
    \begin{subfigure}[b]{0.45\textwidth}
        \includegraphics[trim = 0.5cm 0.2cm 1.5cm 0.5cm, clip, width=\textwidth]{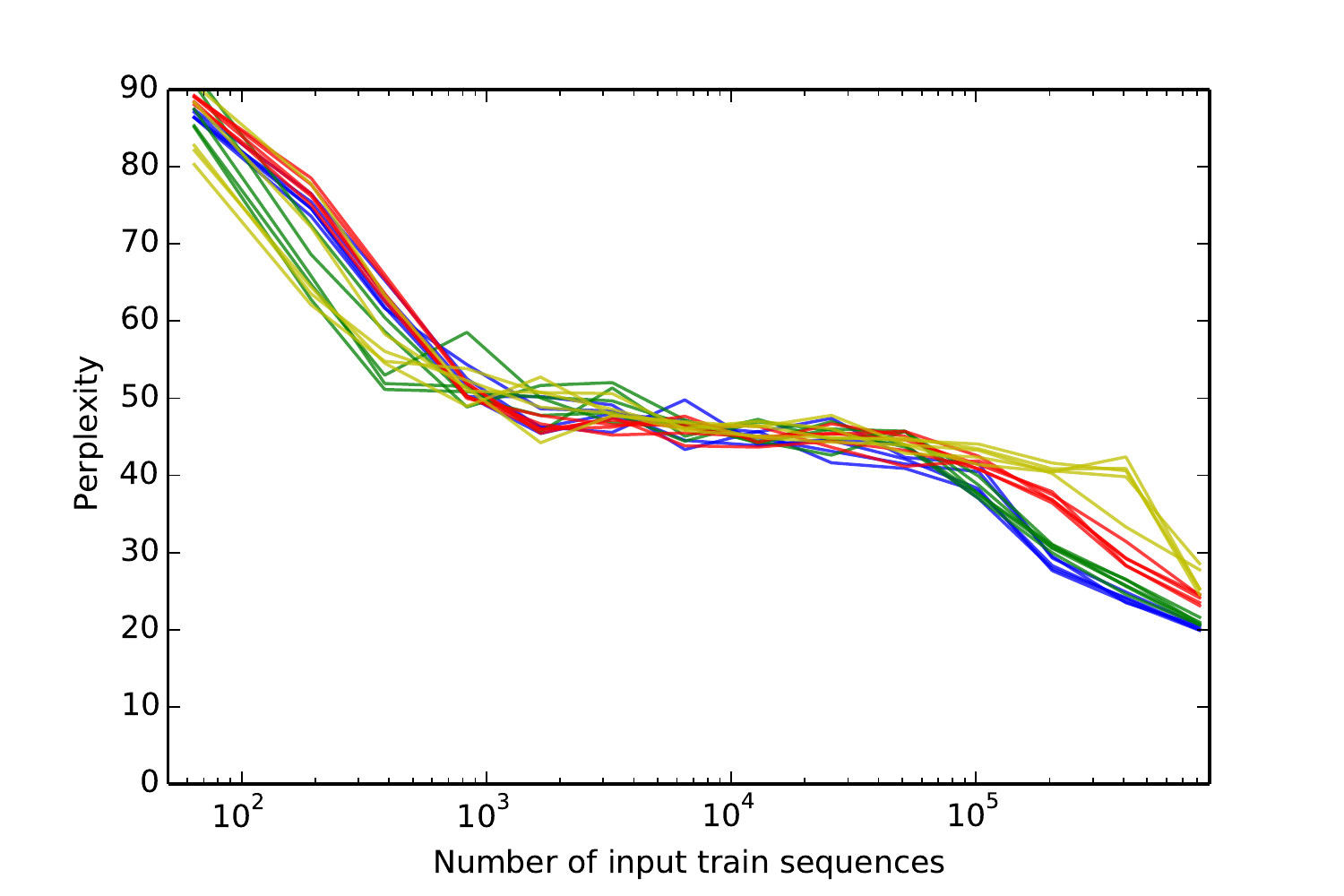}
        \caption{Music}
        \label{fig:music2architectures}
    \end{subfigure}
\end{flushleft}
    \caption{Comparing RNN architectures with scheme 2. Blue: $r=1, \gamma=128$; Green: $r=1, \gamma=512$; Red: $r=2, \gamma=128$; Yellow: $r=2, \gamma=512$.}
    \label{fig:exp2architectures}
\end{figure}

We also perform the same experiments with scheme 2, for which the results are shown in Figure \ref{fig:exp2architectures}.
The same behavior with respect to the architectural differences is observed as in the first scheme.
But now the networks converge somewhat slower, which can be seen especially for the Music dataset by comparing Figures \ref{fig:music1architectures} and \ref{fig:music2architectures}.
On the plus side, the performance curves are smoother than for scheme 1.
Both effects can be explained by the fact that there is only one loss signal at the end of each training sequence, which makes learning slower, but the backpropagated gradient is of higher quality, since more historical characters are taken into account.
From Figures \ref{fig:exp1architectures} and \ref{fig:exp2architectures} we conclude that the RNN architecture indeed influences the efficiency of the training procedure, but that the same effect is observed globally across datasets and training schemes.
The best architecture for all four datasets has parameters $r=1$ and $\gamma=512$, i.e.~the green plots.
This specific architecture will therefore be used in the next experiments.

\begin{figure}[h]
\begin{flushleft}
    \begin{subfigure}[b]{0.45\textwidth}
        \includegraphics[trim = 0.5cm 0.2cm 1.5cm 0.5cm, clip, width=\textwidth]{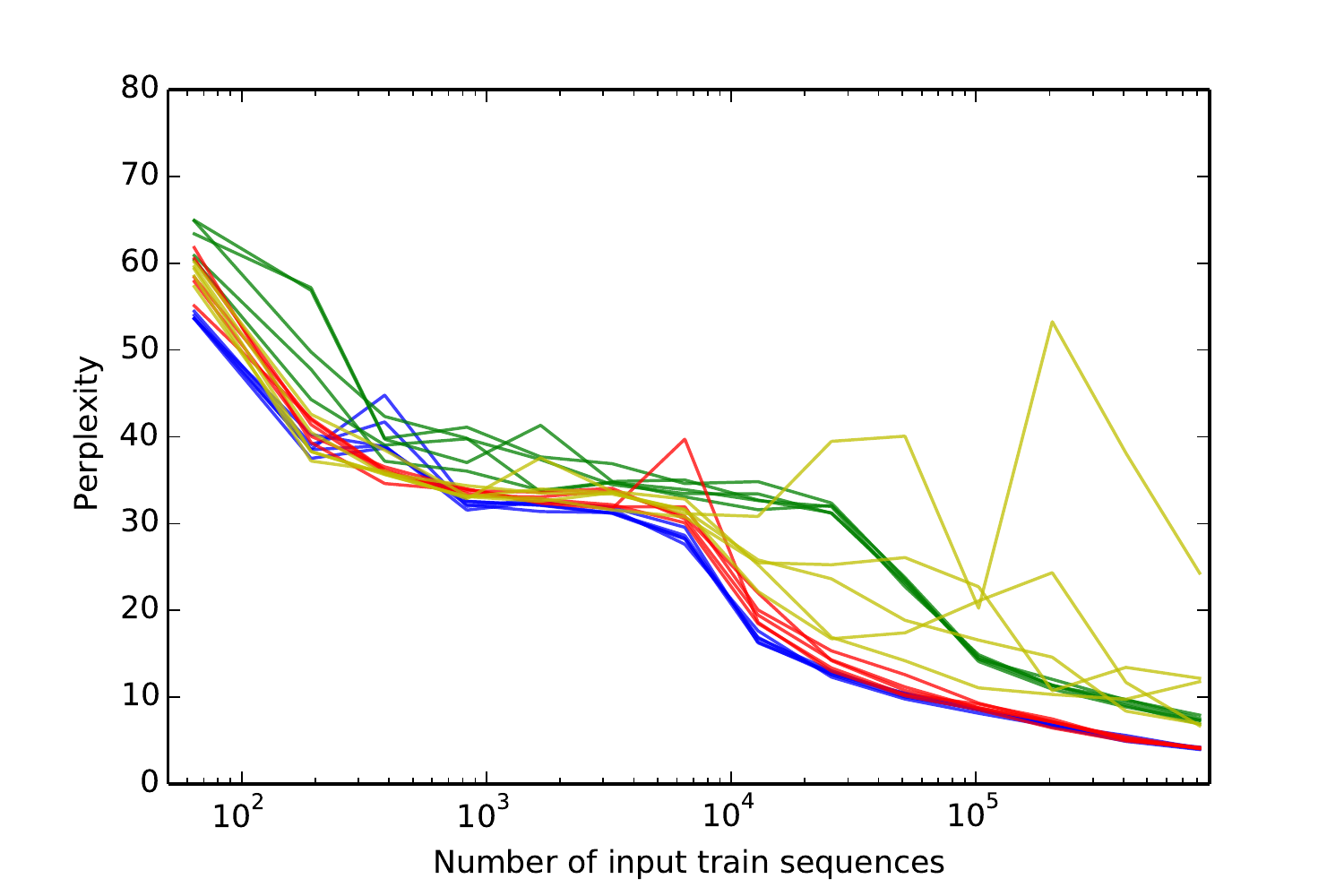}
        \caption{English}
        \label{fig:shakespeareschemes}
    \end{subfigure}
    \begin{subfigure}[b]{0.45\textwidth}
        \includegraphics[trim = 0.5cm 0.2cm 1.5cm 0.5cm, clip, width=\textwidth]{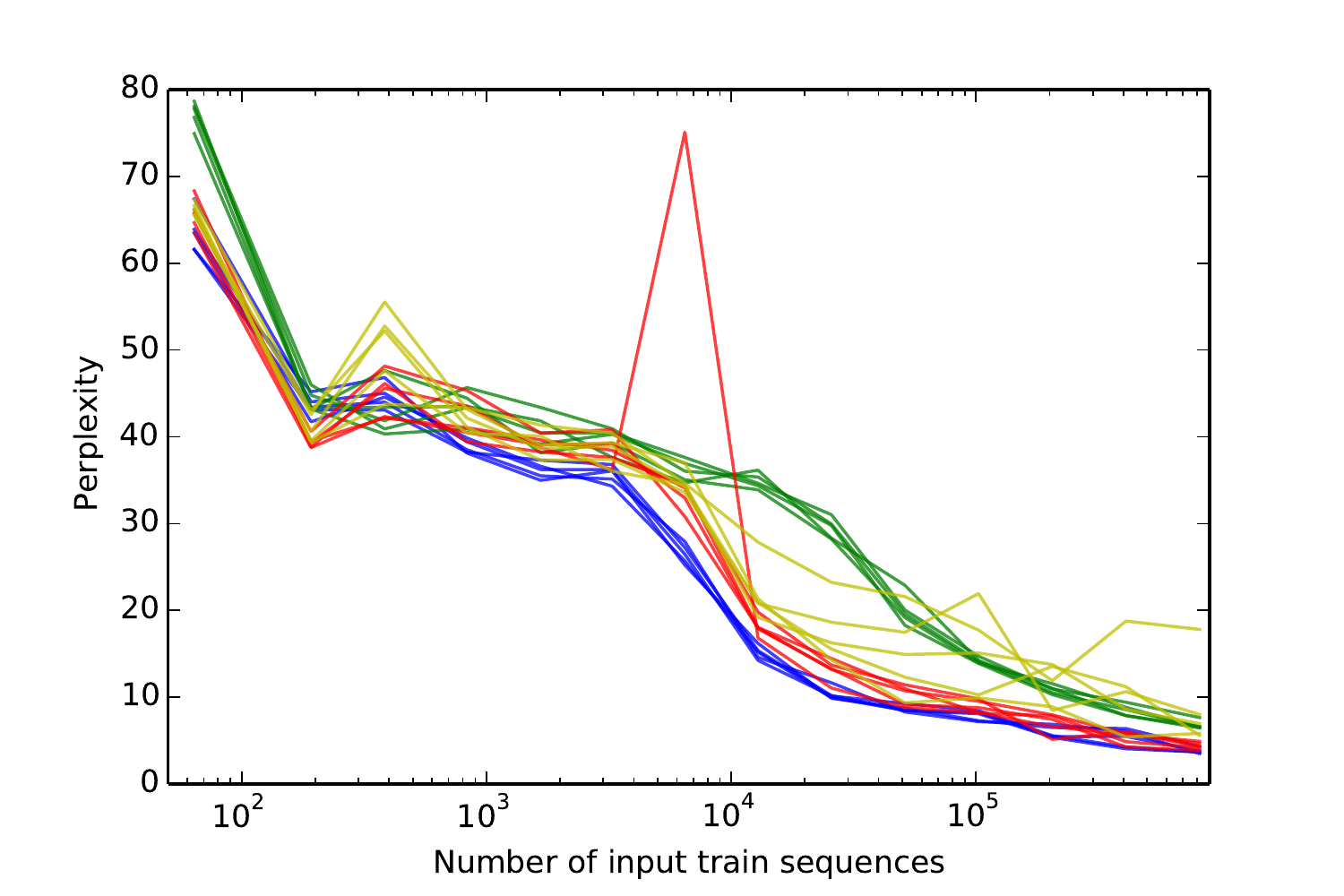}
        \caption{Finnish}
        \label{fig:finnishschemes}
    \end{subfigure}
    \begin{subfigure}[b]{0.45\textwidth}
        \includegraphics[trim = 0.5cm 0.2cm 1.5cm 0.5cm, clip, width=\textwidth]{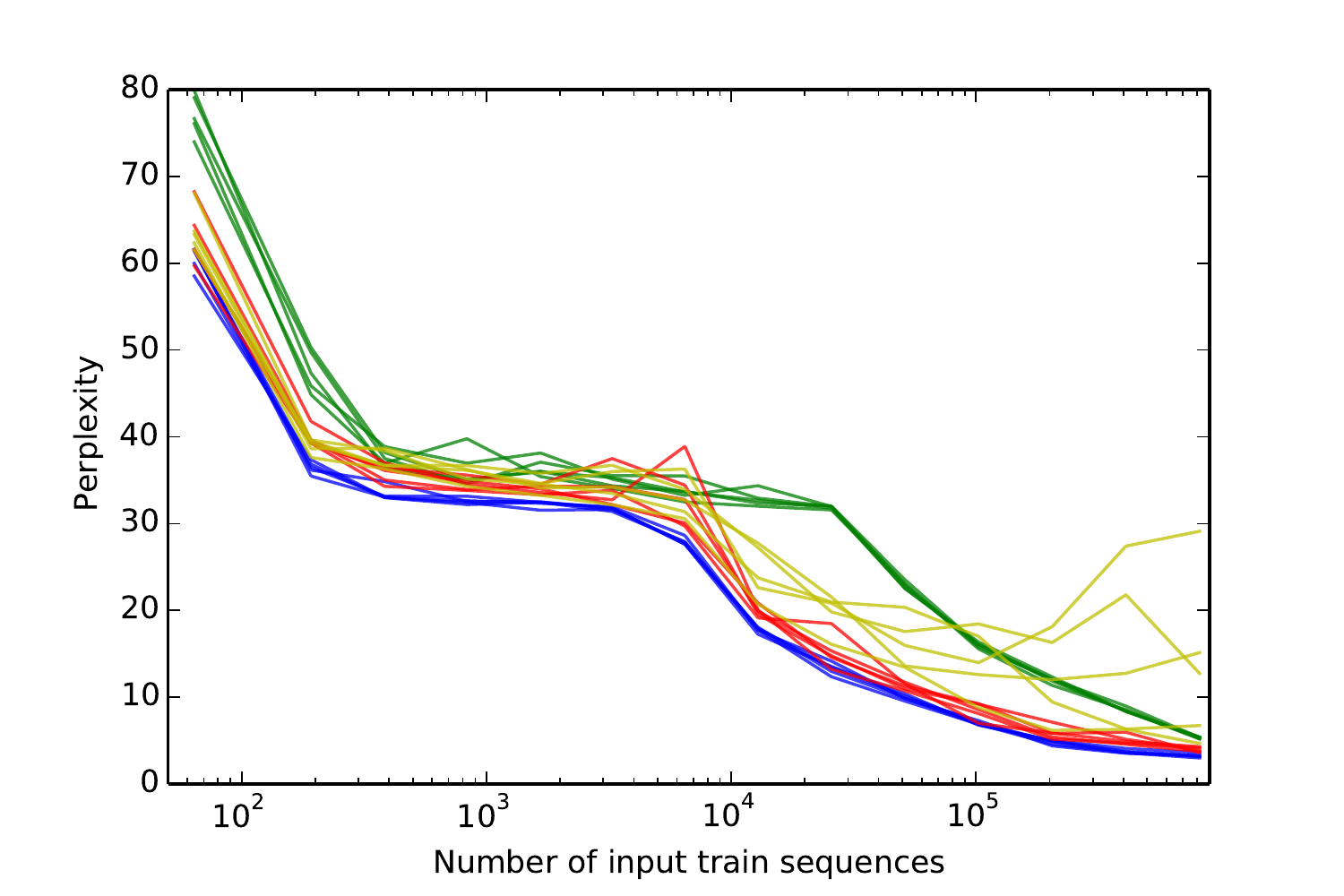}
        \caption{Linux}
        \label{fig:linuxschemes}
    \end{subfigure}
    \begin{subfigure}[b]{0.45\textwidth}
        \includegraphics[trim = 0.5cm 0.2cm 1.5cm 0.5cm, clip, width=\textwidth]{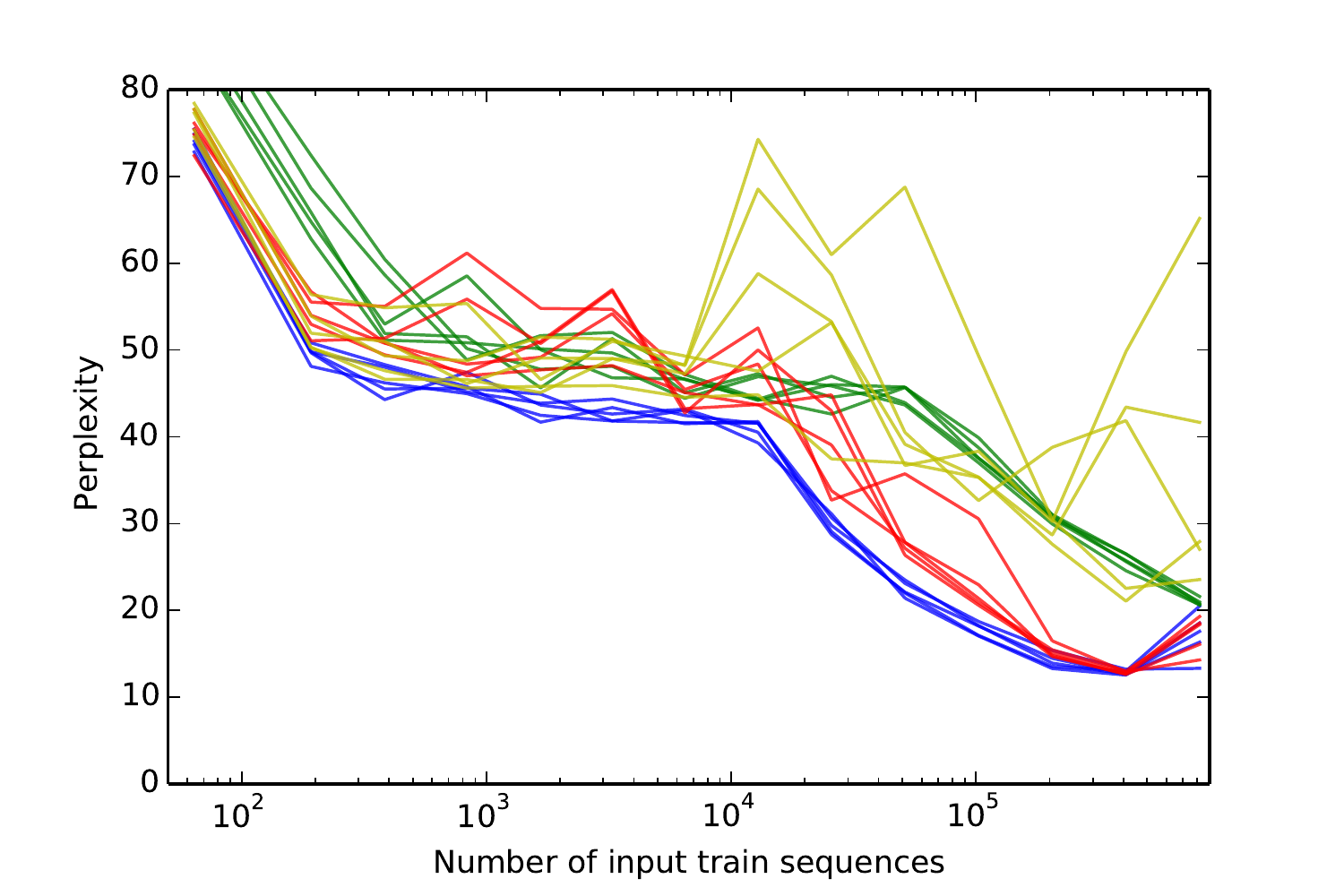}
        \caption{Music}
        \label{fig:musicschemes}
    \end{subfigure}
\end{flushleft}
    \caption{Comparing RNN schemes. Blue: scheme 1; Green: scheme 2; Red: scheme 3; Yellow: scheme 4.}
    \label{fig:expschemes}
\end{figure}

Next, all schemes are compared on the different datasets.
As mentioned above, the architecture with $r=1, \gamma=512$ is used.
The perplexity plots are gathered in Figure \ref{fig:expschemes}.
We see that schemes 1 and 2 are very robust across datasets, but also across different settings of $k_1$, since all lines lie very close to each other.
Scheme 1 is also the best performing in terms of perplexity.
The performance of scheme 2 is overall worse compared to scheme 1, which is probably due to the fact that learning occurs more slowly, as argued before.
The performance of scheme 3 is comparable to the first scheme, but only very slightly worse and robust.
Since the training procedure of schemes 1 and 3 is the same, we hypothesize that the sampling procedure of scheme 3 sometimes has difficulties recovering from errors, which can be carried across many time steps.
Another reason is that the RNN has not learned to make predictions for sequences longer than $k_2$ tokens.
We also mention that for schemes 1, 2 and 3 we experimented with randomly shuffling all training sequences instead of circularly shifting the train set, as explained in Section \ref{sec:experimentalsetup}, but this did not lead to different observations.
In scheme 4 the hidden state is transferred across sequences during training, which appears to solve this problem, at least for some configurations of $k_1$.
All configurations for scheme 4 start with the same performance as for scheme 3, but after around 200 train batches---i.e.~12,800 train sequences in the graph---some configurations start diverging, for which we cannot isolate any consistent motivation or explanation.
From the figures we see that this behavior is also heavily dependent on the dataset; the difference between e.g.~the Finnish and Music dataset is notable.

\begin{figure}[h]
\begin{flushleft}
    \begin{subfigure}[b]{0.45\textwidth}
        \includegraphics[trim = 0.5cm 0.2cm 1.5cm 0.5cm, clip, width=\textwidth]{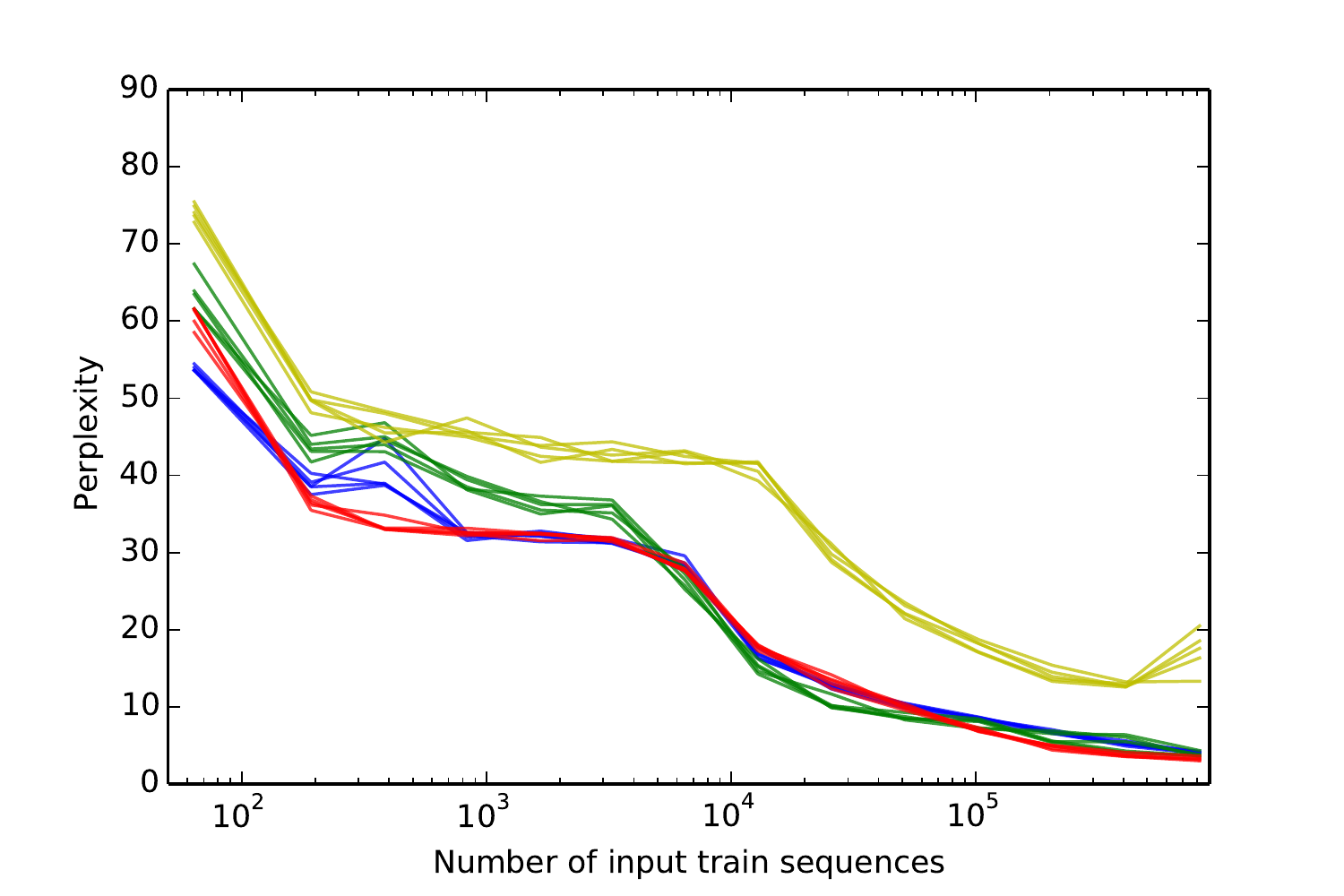}
        \caption{Scheme 1}
        \label{fig:scheme1datasets}
    \end{subfigure}
    \begin{subfigure}[b]{0.45\textwidth}
        \includegraphics[trim = 0.5cm 0.2cm 1.5cm 0.5cm, clip, width=\textwidth]{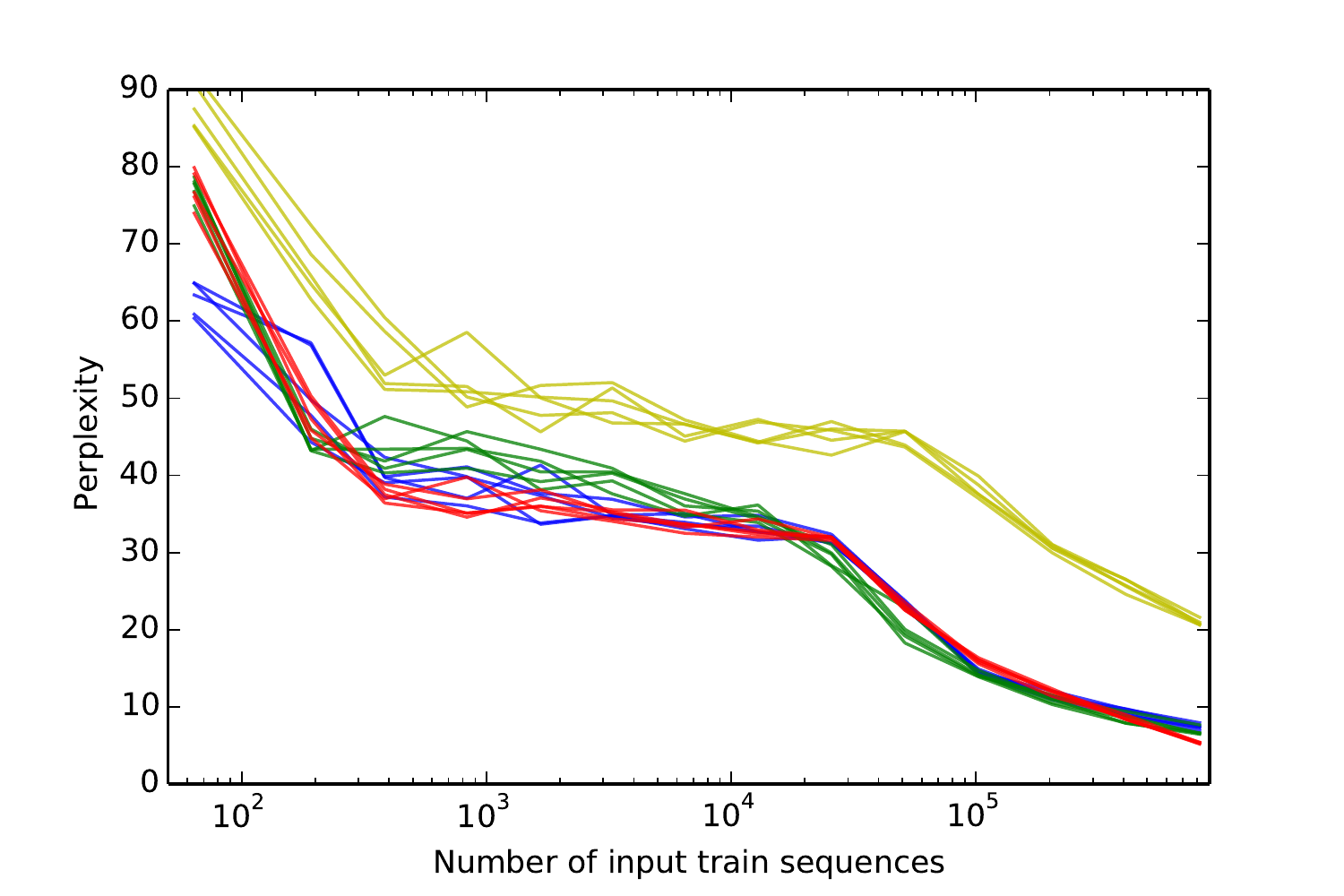}
        \caption{Scheme 2}
        \label{fig:scheme2datasets}
    \end{subfigure}
    \begin{subfigure}[b]{0.45\textwidth}
        \includegraphics[trim = 0.5cm 0.2cm 1.5cm 0.5cm, clip, width=\textwidth]{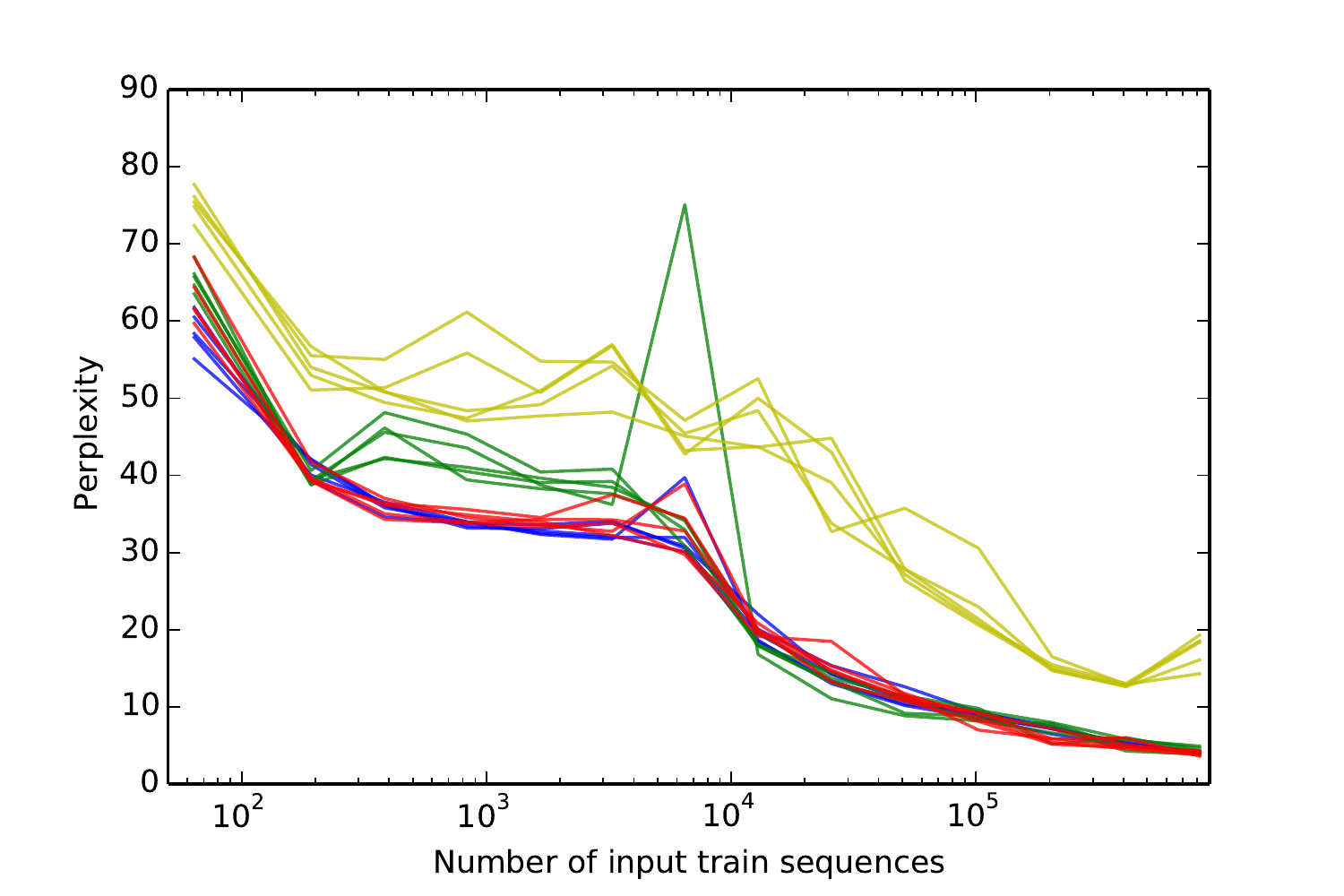}
        \caption{Scheme 3}
        \label{fig:scheme3datasets}
    \end{subfigure}
    \begin{subfigure}[b]{0.45\textwidth}
        \includegraphics[trim = 0.5cm 0.2cm 1.5cm 0.5cm, clip, width=\textwidth]{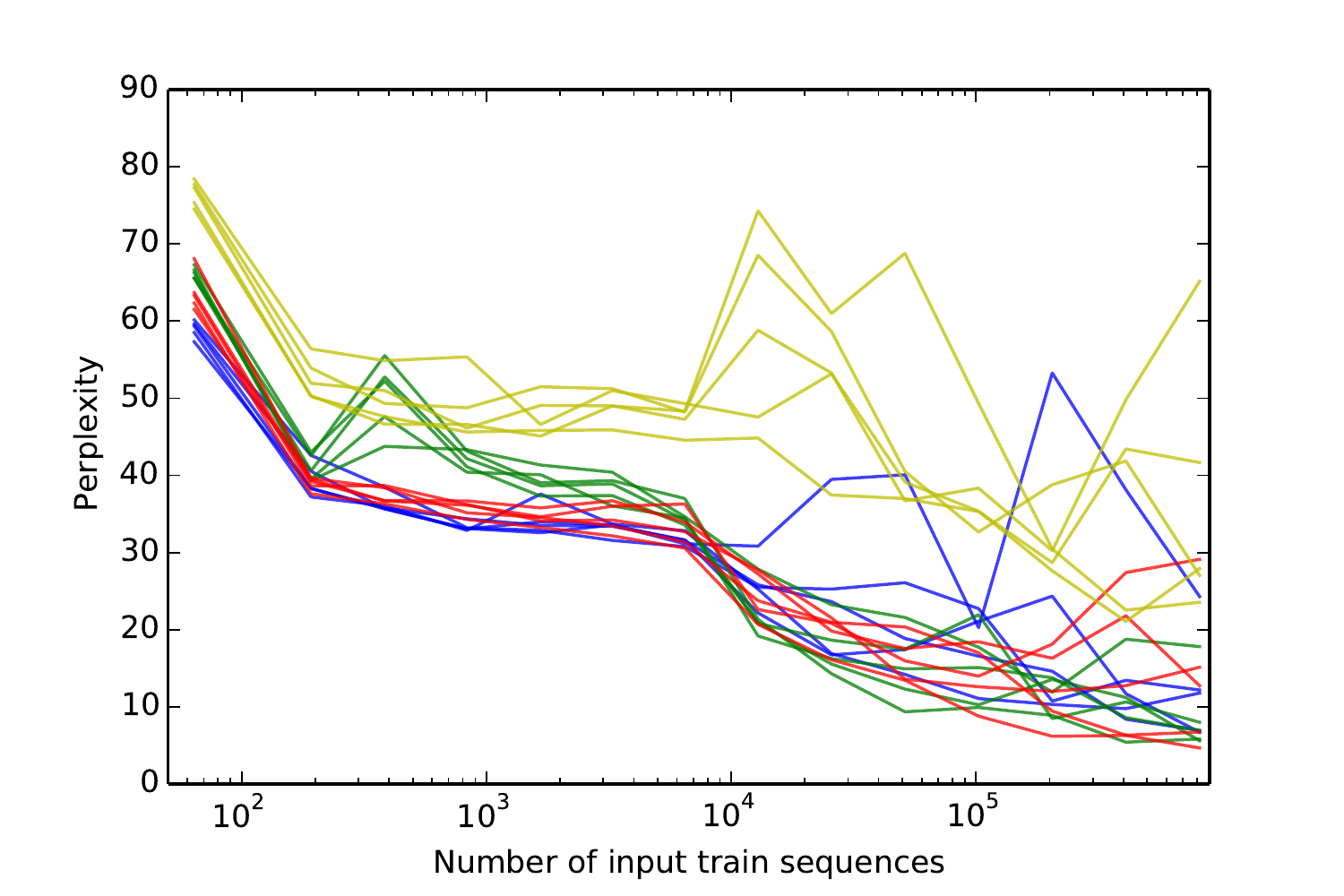}
        \caption{Scheme 4}
        \label{fig:scheme4datasets}
    \end{subfigure}
\end{flushleft}
    \caption{Comparing datasets. Blue: English; Green: Finnish; Red: Linux; Yellow: Music.}
    \label{fig:expdatasets}
\end{figure}

It is also interesting to take a look at a comparison between performances on different datasets for the same scheme.
These curves are plotted in Figure \ref{fig:expdatasets}.
For all schemes we notice that the performance on the English, Finnish and Linux datasets is almost equal; only the Music dataset seems harder to model with the same RNN architecture.
What we also observe is that scheme 1 is very robust against changes in training parameters, since all curves lie very close to each other.
There is more variance in this for scheme 2, even more for scheme 3, and it is highest for scheme 4.

At this point we would also like to discuss data efficiency.
For small values of $k_1$, we use less data at a particular point in the training process compared to larger values of $k_1$.
This is important when data resources are scarce.
From Figure \ref{fig:expdatasets} it is noticeable that, at least for the same scheme, the lines for different values of $k_1$ lie very close to each other.
From these experiments, a general conclusion could be to use a small value of $k_1$ in order to be as data efficient as possible.
The choice of $k_1$, after all, seems to have less impact than the choice of training scheme.
Additionally, using a small value of $k_1$ improves label reuse in the multi-loss training algorithms.
This can approximately be quantified by $\nicefrac{k_2}{k_1}$, i.e.~the number of times a label is reused in the training process.

Up until now we have been comparing the performance of different RNN models and schemes in terms of the number of train sequences used up until a certain point in time.
But the models can also be compared in terms of absolute training and sampling time, which will give us an overview of which configurations are the fastest.
In the next experiment, we calculate the average training time per batch and sampling time for a single token on the English dataset.
We will vary the scheme that we use for training, as well as the RNN architecture.
Concerning the $k_1$ training parameter, there will be almost no difference in training time, so in all measurements we use $k_1=40$.
The numbers are shown in Table \ref{table:timemeasurements}.
It is no surprise that schemes 1, 3 and 4 have almost equal training time per batch, while scheme 2 trains significantly faster since we only need to compute one softmax output for each training sequence.
It is however noticeable that the more complex the RNN architecture, the smaller the relative difference in training time, with a decrease of 25\% for the $r=1, \gamma=128$ architecture and just 9\% for the $r=2, \gamma=512$ architecture.
Regarding the sampling times, we see that the 3rd and 4th schemes are faster by a factor of 10 up to 20 compared to schemes 1 and 2, since there is no need to propagate an entire sequence through the RNN to sample a new token.

\begin{table}[h]
\small
\caption{Absolute training time per batch and sampling time for 1 token using different RNN configurations on the English dataset, $k_1=40$. Measurements in ms. The variance is negligible.}
\label{table:timemeasurements}
\begin{tabular}{l | c c c c }
\toprule
          & $r=1, \gamma=128$ & $r=1, \gamma=512$ & $r=2, \gamma=128$ & $r=2, \gamma=512$\\
\hline
Scheme 1  & 60.5 / 7.0 	& 138.8 / 21.7	& 106.5 / 13.7	& 267.4 / 43.3\\
Scheme 2  & 46.5 / 7.0	& 114.6 / 21.7	& 93.1 / 13.7	& 245.8 / 43.3\\
Scheme 3  & 60.5 / 0.7	& 138.8 / 1.2	& 106.5 / 1.1	& 267.5 / 2.2\\
Scheme 4  & 60.6	 / 0.7  & 138.9 / 1.2	& 106.6 / 1.1	& 267.6 / 2.2\\
\bottomrule
\end{tabular}
\end{table}

\begin{figure}[h]
\begin{flushleft}
    \begin{subfigure}[b]{0.45\textwidth}
        \includegraphics[trim = 0.5cm 0.2cm 1.5cm 0.5cm, clip, width=\textwidth]{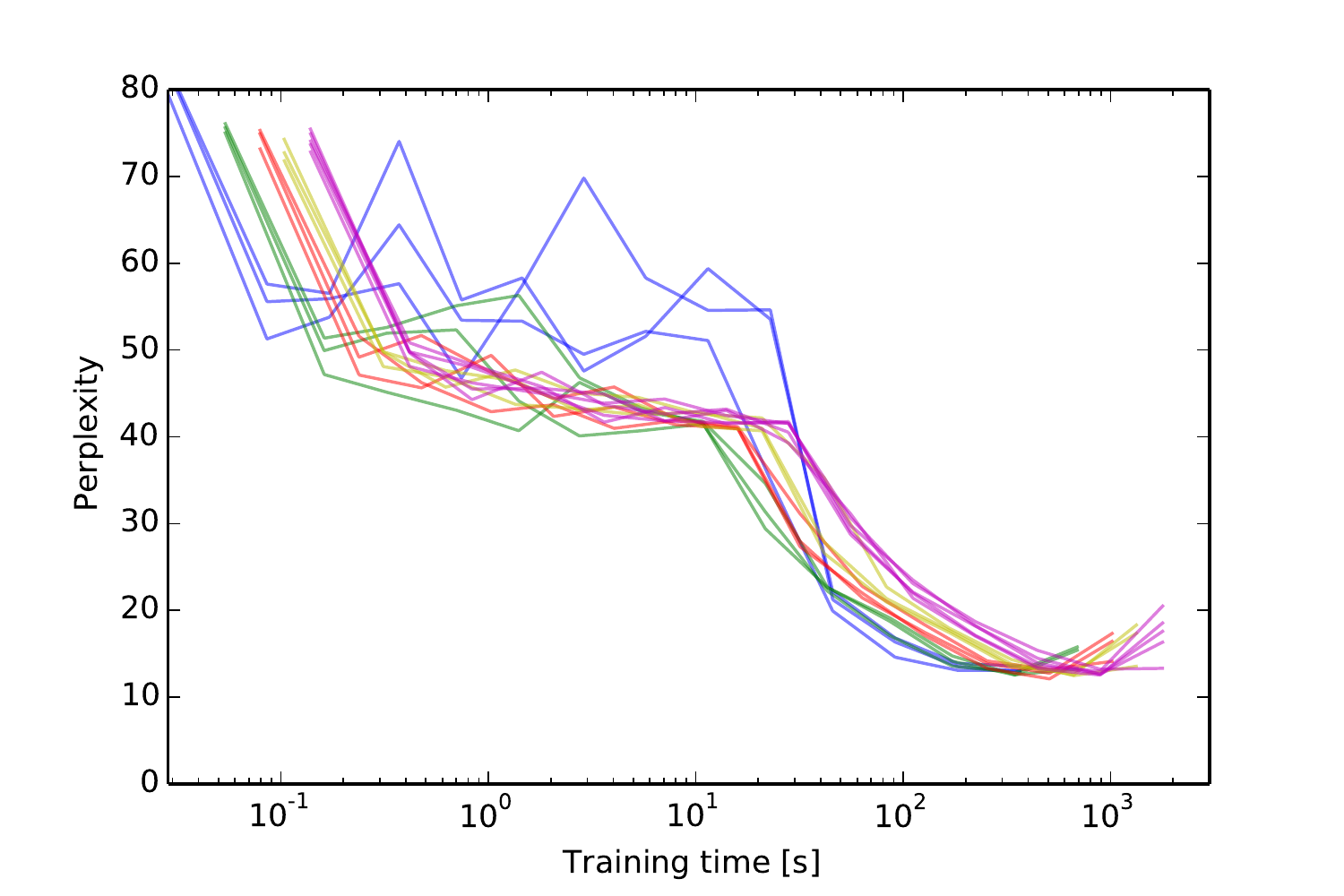}
        \caption{Scheme 1}
        \label{fig:scheme1k2}
    \end{subfigure}
    \begin{subfigure}[b]{0.45\textwidth}
        \includegraphics[trim = 0.5cm 0.2cm 1.5cm 0.5cm, clip, width=\textwidth]{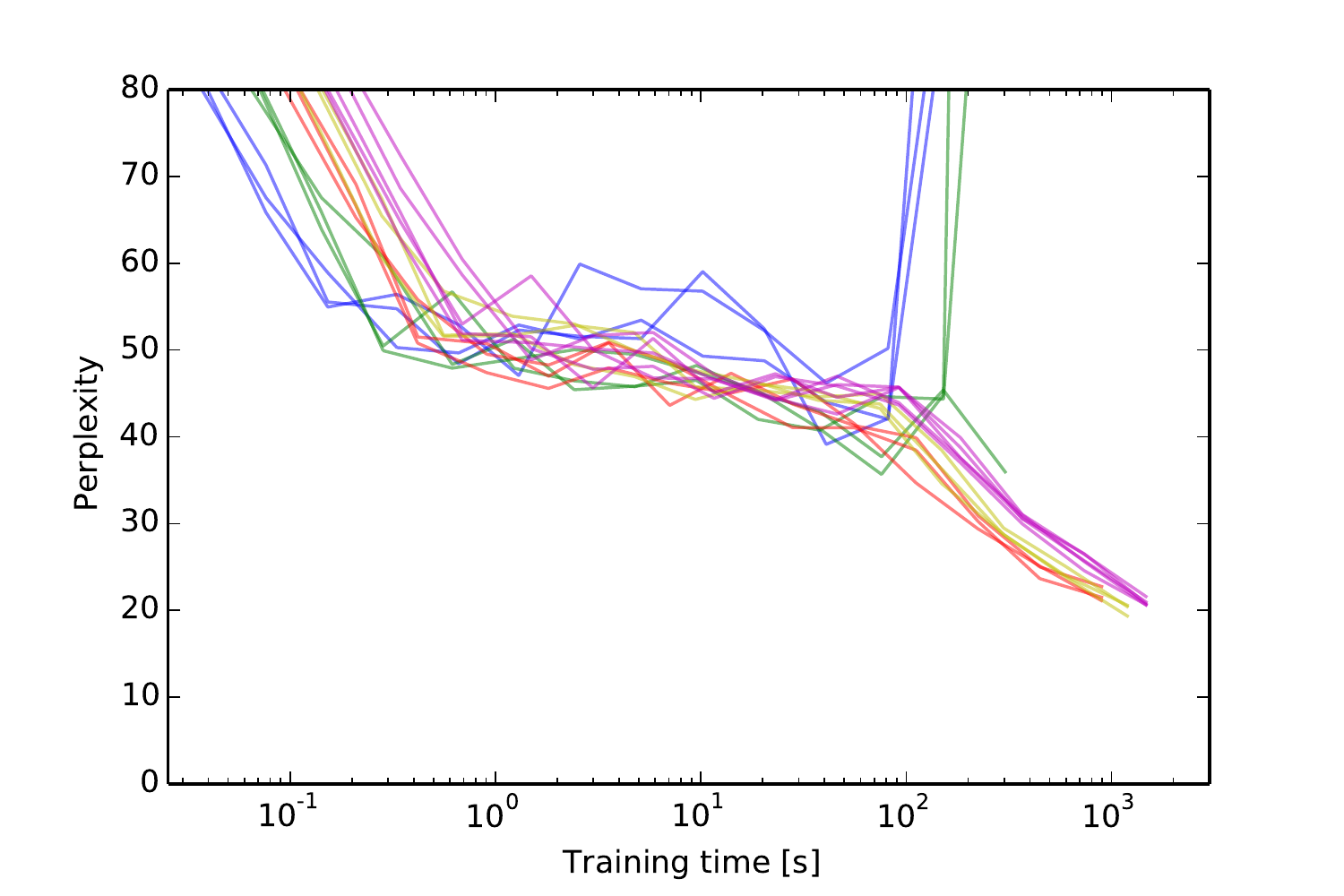}
        \caption{Scheme 2}
        \label{fig:scheme2k2}
    \end{subfigure}
    \begin{subfigure}[b]{0.45\textwidth}
        \includegraphics[trim = 0.5cm 0.2cm 1.5cm 0.5cm, clip, width=\textwidth]{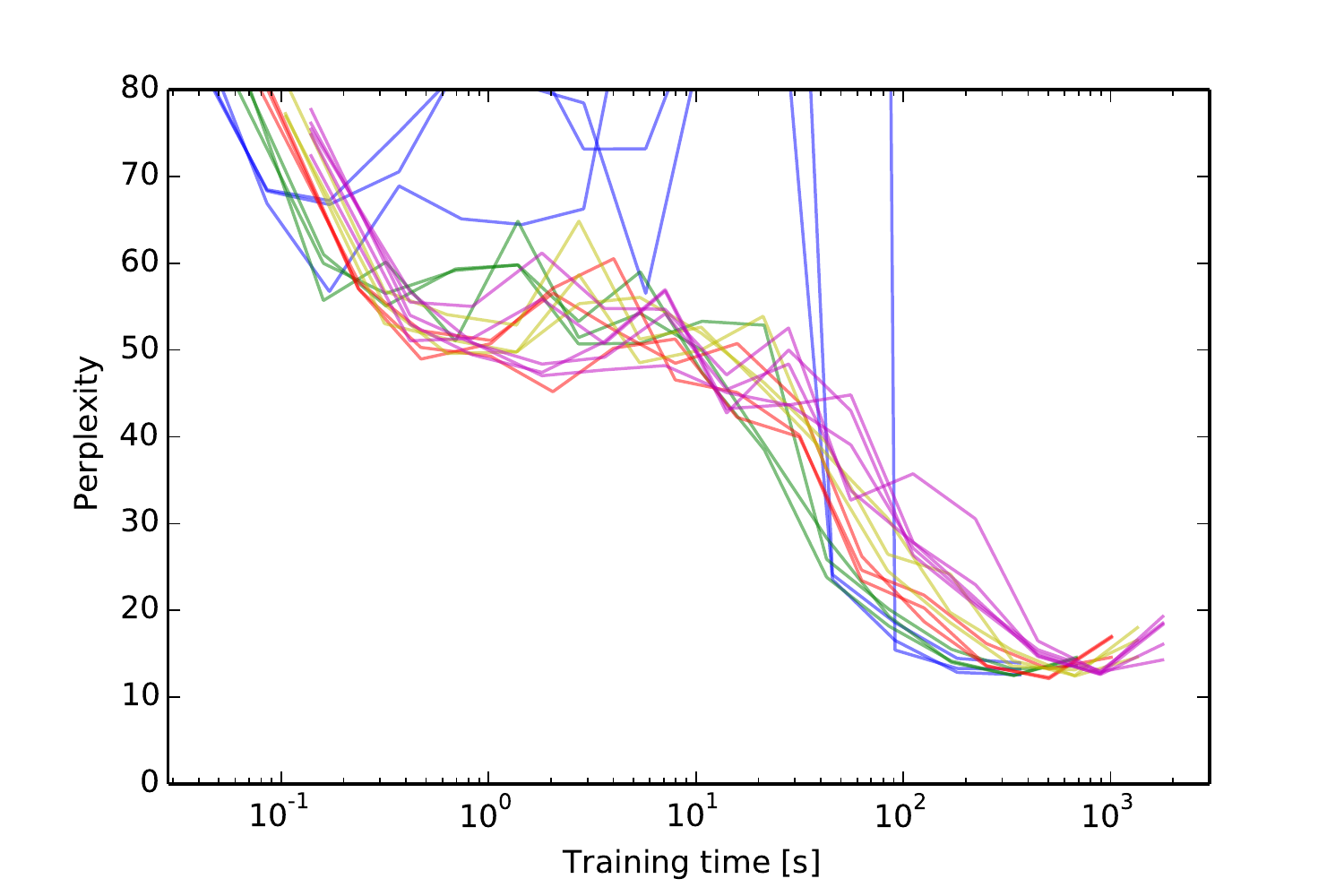}
        \caption{Scheme 3}
        \label{fig:scheme3k2}
    \end{subfigure}
    \begin{subfigure}[b]{0.45\textwidth}
        \includegraphics[trim = 0.5cm 0.2cm 1.5cm 0.5cm, clip, width=\textwidth]{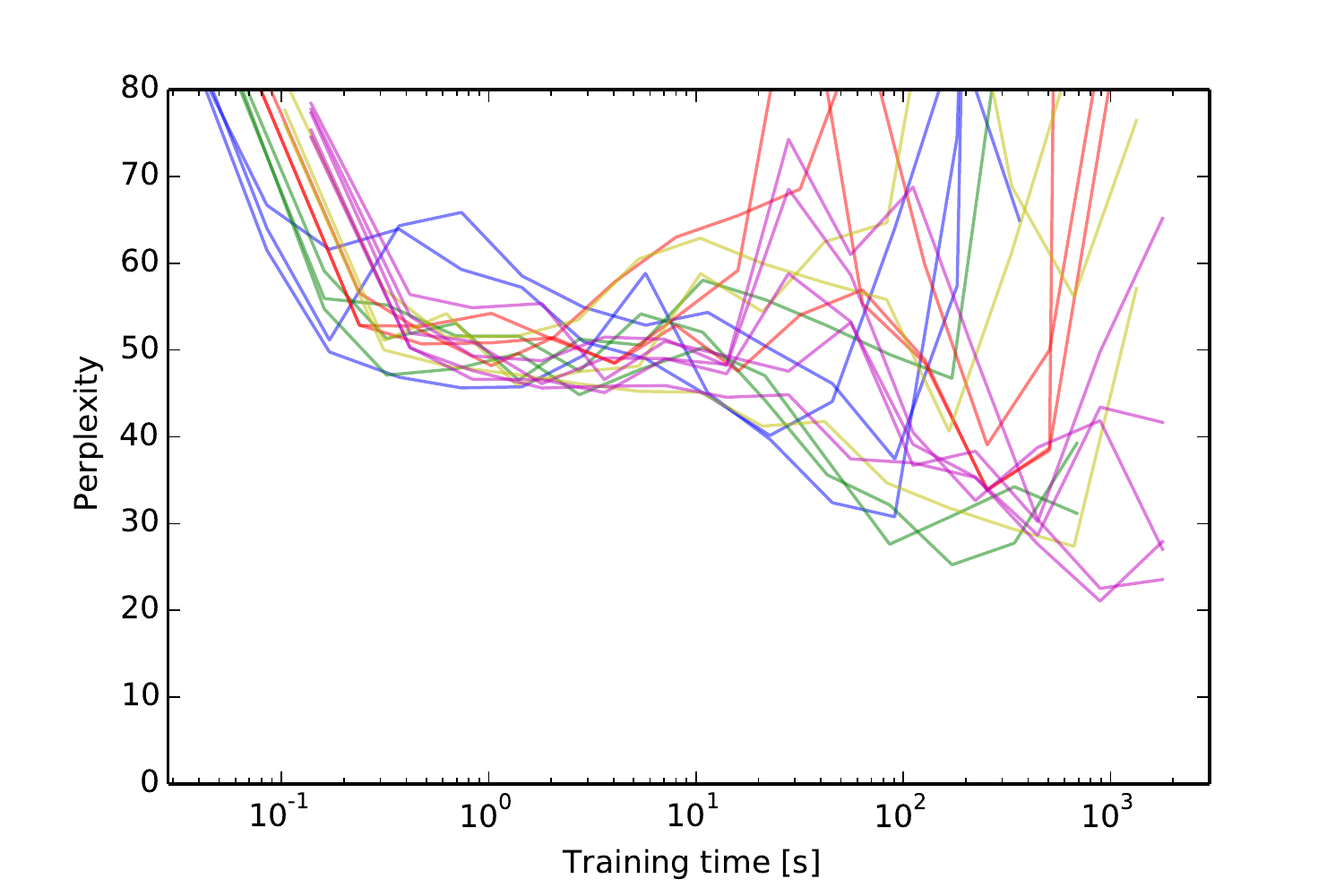}
        \caption{Scheme 4}
        \label{fig:scheme4k2}
    \end{subfigure}
\end{flushleft}
    \caption{Performance with respect to training time by comparing $k_2$ values on the Music dataset. Blue: $k_2 = 20$; Green: $k_2 = 40$; Red: $k_2 = 60$; Yellow: $k_2 = 80$; Magenta: $k_2=100$.}
    \label{fig:expk2}
\end{figure}

We also compare the performance of the different schemes with respect to changes in the $k_2$ parameter.
For each scheme we perform five experiments, for which $k_2$ is set successively to 20, 40, 60, 80 and 100.
After setting $k_2$, the $k_1$ parameter is set to $k_2$, $\nicefrac{2k_2}{3}$ and $\nicefrac{k_2}{3}$, rounded to the nearest integer.
Every experiment is performed on the Music dataset, since, based on previous experiments, we expect to gain most insights on it.
We report the model perplexity on the test set as a function of the elapsed training time, and we train again for a total of 12,800 batches.
The results are shown in Figure \ref{fig:expk2}, in which the $y$-axis is clipped to a maximum of 80 to achieve the most informative view.
We see that the smaller the $k_2$ value, the faster we have trained all batches, since it leads to a shorter BPTT.
The first scheme is again the most robust against a changes in $k_2$.
Only the shortest sequence lengths behave more noisily in the first 10 seconds of training, but all configurations are able to reach a similar optimal perplexity.
The second scheme trains much slower than scheme 1, and experiences instability problems for small sequence lengths of 20 and 40.
The configurations with $k_2 \geq 60$ are all very stable, but have not yet fully converged after 12,800 batches.
For scheme 3 we see almost the same behavior as in scheme 1, with all configurations reaching the same optimal perplexity.
But, just as we saw before, the robustness against changes in $k_1$ is worse.
This is especially true for small values of $k_2$, as shown by the blue lines in Figure \ref{fig:scheme3k2}.
Finally, for scheme 4 we see that almost all configurations are unstable and behave very noisily.
Two configurations with $k_2=20$ even achieve a final perplexity of around 350; lowering $k_1$ for small values of $k_2$ seems to help in this case.

We conclude this experimentation section with a few recommendations.
We found that the global behavior of the different schemes is nearly independent of the used dataset.
This is good news, since we do not have to tune the learning and sampling procedure to the dataset at hand.
In this respect, we arrive at the following conclusions:

\begin{itemize}
\item 
In terms of training schemes, the multi-loss approach (scheme 1 and 3) is recommended. Compared to the single-loss approach (scheme 2), multi-loss training is more efficient. The faster individual iterations of the single-loss approach cannot compensate for the benefit of combining the loss over multiple positions in the sequence, when considering the total train time.
\item 
Our general recommendation is to avoid training procedures in which the hidden state is transferred between input sequences (scheme 4). Training is as efficient as the multi-loss approach without transferred hidden states (scheme 3), but less robust. 
On noisy datasets, such as the Music dataset in our experiments, transferring hidden states is likely to cause an unstable behavior.
\item
On the sampling side, there is a trade-off between windowed sampling and progressive sampling. By comparing scheme 1 and 3, it is seen that windowed sampling is more robust than progressive sampling. However, the latter is more efficient by construction, as it samples the next character based on the current one and the hidden state, instead of each time performing a forward pass over a (possibly long) sequence as in the windowed sampling approach.
\end{itemize}

\section{Future research tracks}
\label{sec:futurework}
We include one final section on future research tracks in the area of training and sampling procedures for character-level RNNs.
In this paper we have made an attempt at isolating the four most common schemes that have been or are being used in literature.
There are however multiple hybrid combinations that can be identified and investigated in the future.
The most straightforward extension is an intermediate form between single- and multi-loss training.
For example, an extra parameter $k_3$ could be identified, for which $k_3 \leq k_2$, that defines the number of time steps for which the loss is calculated and aggregated. 
The edge cases $k_3 = 1$ and $k_3 = k_2$ correspond respectively to the single-loss and multi-loss training procedures.
One other possibility is to decay the loss at each time step (linearly or exponentially) and combine these through a linear combination to calculate the final loss.
For a single training sequence $s$ this results in:
\begin{align*}
\texttt{loss}_s = \sum_{(x_i, x_{i+1}) \in s} \mathcal{L}\left( \mathcal{R}(x_i), x_{i+1}\right)\cdot \gamma_i,
\end{align*}
with $\gamma_i = \nicefrac{i}{\mathrm{length}(s) - 1}$ or $\gamma_i = \exp(i - \mathrm{length}(s) + 1)$ for resp.~linear and exponential decay.
Consequentially, the resulting gradient is scaled similarly, thereby reducing the contribution of the first few tokens in the sequence to the total loss.

\section{Conclusion}
\label{sec:conclusion}
We explained the concept of character-level RNNs and how such models are typically trained using truncated backpropagation through time.
We then introduced four schemes to train character-level RNNs and how to sample new tokens from such models.
These schemes differ in how they approximate the truncated backpropagation through time paradigm: how the RNN outputs are combined in the final loss, and whether the hidden state of the RNN is remembered or reset for each new input sequence.
After that, we evaluated each scheme against different datasets and RNN configurations in terms of predictive performance and training time.
We showed that our conclusions remain valid across all these different experimental settings.

Perhaps the most surprising result of the study is that conditional multi-loss training, in which the hidden state is carried across training sequences, often leads to unstable training behavior depending on the dataset.
This contrasts sharply with the observation that this training procedure is used most often in literature, although it requires meticulous bookkeeping of the hidden state and a carefully designed batching method.
%
Single-loss training is, compared to multi-loss, slower regarding the number of used train sequences.
An advantage of single-loss training, however, is that we encourage the network to make predictions on a long-term basis, since we only backpropagate one loss defined at the end of a sequence.

We saw that progressive sampling is slightly less robust to changes in training parameters compared to windowed sampling, especially for datasets that are more difficult to model, as we showed with the Music dataset. The main advantage of progressive sampling is that it is orders of magnitudes faster than windowed sampling.

\section*{Conflicts of interest}
Funding: the hardware used to perform the experiments in this paper was funded by Nvidia.\\
Conflict of interest: Cedric De Boom is funded by a PhD grant of the Research Foundation - Flanders (FWO). The other authors declare that they have no conflicts of interest.

\bibliographystyle{spmpsci}      

\end{document}